# Healthcare cost prediction for heterogeneous patient profiles using deep learning models with administrative claims data

Mohammad Amin Morid[1], Olivia R. Liu Sheng[2]

[1] Department of Information Systems and Analytics,
Leavey School of Business,
Santa Clara University, Santa Clara, CA, USA.
500 El Camino Real, Santa Clara, CA, 95053-0382
Phone: +1(408) 554-4629
Email: mmorid@scu.edu

[2] Department of Information Systems, W. P. Carey
School of Business,
Arizona State University, Tempe, AZ, USA.
400 E Lemon St, Tempe, AZ 85281
Phone: +1(801) 792-2248
E-mail: olivia.liu.sheng@asu.edu

## Abstract

Accurate and fair patient cost predictions, which can lead to healthcare payer cost savings, are essential to support effective decision-making regarding health management policies and resource allocations. Patient cost prediction models utilize administrative claims (AC) data collected from multiple healthcare providers, which payers (e.g., government agencies and private insurance companies) rely on for various reimbursement purposes. Both the variety of patient clinical profiles and the multi-source nature of the big data from ACs introduce heterogeneity, which undermines both the generalization power and the algorithmic fairness of cost prediction models. In particular, the prediction performance and economic outcomes—such as both underpayments and overpayments—of these models for high-need (HN) patients with multiple and complex chronic conditions differ from those of healthy patients, as their underlying heterogeneous medical profiles are distinct.

This study, grounded in socio-technical considerations for patient cost prediction, presents two key design insights. First, we designed a channel-wise deep learning framework to reduce AC data heterogeneity through effective representation learning, with a separate channel for each type of code as well as each type of cost. Second, we incorporated humanistic outcomes and a multi-channel entropy measurement into a flexible evaluation design for patient heterogeneity.

We evaluate the effectiveness of the proposed channel-wise framework both internally and externally using two real-world datasets containing approximately 111,000 and 134,000 individuals, respectively. On average, channel-wise models substantially reduce prediction errors by 23% compared to the most competitive single-channel counterparts, leading to respective reductions of 16.4% and 19.3% in overpayments and underpayments for patients. The reduction in bias for predictions involving HN patients is more significant than for other patient groups. Our findings offer important implications for decision-makers in healthcare and other fields facing similar socio-technical challenges related to the interplay between diverse population behaviors and data heterogeneity.

*Keywords: High-need patients, Heterogeneity, Cost prediction, Risk adjustment model, Representation learning, Channel-wise deep learning*

# 1. Introduction

Rising healthcare costs in the United States have become a critical and pressing issue for consumers, providers, payers (e.g., health plan providers, Medicare, and Medicaid), and governmental financiers of healthcare alike (Duncan et al. 2016, Martin et al. 2019). In order to reduce patient expenses while also improving access to resources, healthcare organizations and government agencies, such as the Centers for Medicare & Medicaid Services (CMS), are actively engaging in cost analysis and prediction efforts. Since the early 2000s, healthcare insurance providers have attempted to move from the fee-for-service payment model, which incentivizes the over-provision of services and thereby increases healthcare costs, to a fee-for-value (FFV) model, where payments are based on the quality of provided care rather than the volume of services performed (Cucciare and O'Donohue 2006).

Capitation payment (CP) is one of the most common models for implementing FFV, in which healthcare providers or health plans are paid a fixed amount per patient to cover the cost of providing healthcare services—including preventive care, routine office visits, and any necessary medical treatments (Volk et al. 2019). For example, in some Medicare insurance settings, such as Medicare Advantage (MA) plans, CMS is the payer side of the CP, and the payees are the health plans (e.g., MA plans) offered by private insurance companies, who can then pay the healthcare providers through CP or non-CP payment models in order to cover the patients' costs (Noyes et al. 2008). In contrast, in some commercial insurance settings, health maintenance organizations, which are operated by private insurance companies, pay medical groups (i.e., healthcare providers) to cover such costs (Robinson and Casalino 1995). It is important to note that a variety of CP and non-CP payment models may exist in the healthcare system's supply chain, with each entity acting as a payer on one end and a payee on the other. Given that this study's experimental setting centers on Medicare patients, we refer to CMS as the payer and health plans as the payees throughout this paper.

Around 5% of the U.S. population accounts for approximately 50% of the nation's healthcare spending (Bélanger et al. 2019). Since patients' differential needs and health conditions can impact the cost of care provided by patients' healthcare providers in CP models, risk adjustment models (RAMs) are used to reflect the expected (i.e., predicted) annual costs of individual enrollees (Shen and Ellis 2002). By using statistical methods to adjust the CP amount based on the health needs of the serviced patient population, RAMs ensure that healthcare providers are fairly compensated for the care they provide. For instance, CMS commonly uses RAMs in MA plans to determine the fixed amount of total payments for each member in advance of the upcoming year based on the estimated (i.e., expected) annual cost of the member's healthcare needs (Durfey et al. 2018).

While RAMs encourage efficiency and cost containment, such systems have their own deficiencies, including issues with both underpayment and overpayment (Shen and Ellis 2002). Underpayment occurs when the CP amount is set too low and health plans lack enough resources to provide high-quality care to their patients. For example, researchers have found that institutions such as Medicare underpay health plan providers for high-need (HN) patients, including those with multiple and complex-chronic conditions (Frogner et al. 2011). This can lead healthcare providers to avoid high-risk or complex patients with heterogeneous patient profiles, as these patients may be more expensive to treat. Underpayment can also result in health plans cutting corners, avoiding necessary tests or procedures, or otherwise failing to provide adequate care (Park et al. 2017, Van et al. 2016). In contrast, overpayment occurs when the CP amount is set too high, which again incentivizes health plans to cherry-pick healthy, low-risk enrollees, as they may be less expensive to treat and can therefore result in higher profits (Hellander, Woolhandler, et al. 2013, Jason Brown et al. 2014). For instance, researchers have found that Medicare overpays health plan providers for healthy patients that have below-average costs (Pope et al. 2004). Consequently, accurate expected cost prediction for heterogeneous patient profiles can significantly help payers guarantee that health plans receive proper compensation per patient (Kautter et al. 2012). This should reduce the incentives leading to the biased selection of low-risk patients, and it may even create incentives to attract HN patients, because there may be more opportunities for health plans to provide care efficiently and thus generate profits (Schone and Brown 2013).

Cost prediction models designed to aid RAMs are developed using administrative claims (AC) data collected by insurers from healthcare providers' revenue cycle management systems (Min et al. 2019, Sha and Wang 2017). However, the AC data collection process faces several limitations, posing significant challenges to cost predictions with an increasing degree of semantic and relational heterogeneity in AC data as a result of the more complex medical journeys of HN patients (Liu et al. 2023, Luijken et al. 2019, Thomas et al. 2020). Due to its ability to capture and extract complex relationships from raw temporal data (da Silva et al. 2022), the representation learning potential of deep learning models (Wenzhong Guo et al. 2019) offers a promising direction for designing a patient cost prediction framework. Extant studies on cost prediction methods (Bertsimas et al. 2008, Morid et al. 2017) and other deep learning models for predictive non-cost outcomes have not adequately addressed the challenges arising from heterogeneity in AC data for HN patients whose underlying heterogeneous medical needs and complex healthcare journeys are different from those of relatively healthy patients. HN patients may frequently travel long distances to the nearest hospitals in order to see different providers located in multiple clinics in the same day (Byrne et al. 2021, Dharmadhikari and Zhang 2013). For instance, the same HN patient could see an oncologist for colon cancer treatment, a gastrointestinal specialist for information about the side effects of medication, a physical therapist for post-surgery recovery, and a clinical social worker for anxiety and depression—all in one day (Byrne et al. 2021, Dharmadhikari and Zhang 2013). These multiple visits on the same day create a significant amount of heterogeneity in AC data, manifested in the wide variety of medical codes, as well as multiple insurance claims, associated with each HN patient per day submitted by different providers (which may not be related to each other),[1] accentuating the challenges of cost prediction for HN patients.

In this study, we consider relevant social factors in mapping the heterogeneous complexities of patient journeys into various manifestations of AC data heterogeneity—including heterogeneous medical code and cost types, which further encompass the heterogeneity of unknown inter-visit relationships, and orderless codes in the same claim. This mapping fosters the unique integration of a properly contextualized channel-wise deep learning architecture with other synergistic representation learning designs in the proposed prediction framework in order to both tackle the targeted issues of data heterogeneity and to significantly improve the prediction performance and downstream impact for HN patients (Joynt et al. 2017). Concerned by the obstacles to codifying patient need levels according to sophisticated expert knowledge on the severities of health conditions for model evaluation purposes, we devised a flexible and scalable alternative scheme for stratifying patients using a multi-source entropy index for measuring individual-level data heterogeneity. Similar to the evaluation findings based on patient severity heterogeneity, individuals in high entropy strata benefit from a more significant reduction in both prediction errors and payment discrepancy.

Rooted in the socio-technical connections for design-oriented research on patient cost prediction, this study offers two salient design insights according to the abstraction spectrum introduced by (Abbasi et al. 2024). The first design insight extends the cost prediction and design research literature with a new deep learning pipeline of super-additive data heterogeneity reduction mechanisms to effectively address the heightened challenges that arise from HN patient profiles in AC data. The second design insight fills the research gap in investigating patient heterogeneity in prediction performance and impact evaluation design, empowering this research with a flexible, alternative scheme for delineating heterogeneous patient subpopulations. While there is a rich stream of information systems (IS) literature on the taxonomies of evaluation design methods—such as efficacy, accuracy, and economic-operational utility (Prat et al. 2015)—this study suggests considering humanistic outcomes (i.e., healthcare disparities leading to overpayment or underpayment) in the prediction evaluation design. These insights provide valuable implications for decision-makers in healthcare and other problem domains with similar socio-technical considerations underlying the link between heterogeneous population behaviors and data heterogeneity.

[1] For instance, the average numbers of providers, clinical codes and claims for the HN patients who constitute 44% of our main dataset are three times higher than that of patients who are relatively healthy during a day or a longer duration.

# 2. Background

## 2.1. Cost prediction with administrative claims data

**2.1.1. Data generation process for administrative claims data.** While there are various types of healthcare data, patient cost prediction has solely relied on AC data in the literature for two reasons (Morid et al. 2022). First, the most important stakeholders of cost prediction methods are the payers who collect and maintain AC data only with the medical claims they receive from healthcare providers for billing purposes. Second, analyzing patients' costs requires a complete history of patients' service encounters with various healthcare providers, and AC data is the only repository of this multisource information from different providers. ACs consist of two types: medical and pharmacy. The former are provided by caregivers, such as hospitals, and include diagnosis codes, procedure codes, and cost information, while the latter include medication codes and cost information as provided by pharmacies. Used for insurance reimbursement purposes, a payee (e.g., a hospital or pharmacy) may take the medical codes and cost data associated with multiple visits for the same patient over one or multiple days and aggregate them into a single insurance claim, which is time stamped at the day level for submission for reimbursement.

To explain the data collection process for AC data, Figure 1.a. depicts an example of the health service journey of patient *pt1*, involving medical providers (*prov1* and *prov2*) and a pharmacy provider (*prov3*) over six visits, some occurring on the same day. On *d1* the patient visits *prov1*, where she is diagnosed with three diagnosis codes (*DX1*, *DX2*, and *DX3*), takes a lab test (*lab1*, which is captured as *PX4* later), and receives a medication (*RX1*, which is captured as *PX5* later). On *d2*, which can be close to or far from *d1*, the patient has two medical appointments with *prov1* and *prov2*, where she takes a lab test (*lab2*, which is later captured as *PX6*) and goes through two procedures (*PX1* and *PX2*). These two visits may be related to each other (e.g., *prov1* may suggest the appointment with *prov2*), or they may be unrelated (e.g., the patient simply scheduled these two appointments on the same day). Such relationships can only be extracted by accessing the physician notes from *prov1* and *prov2*. After these two visits, the patient goes to a pharmacy (*prov3*) and picks up some medications (*RX1* and *RX2*), which may or may not be related to the two visits she had with *prov1* and *prov2* on that same day. Such relationships can only be extracted by accessing the prescription notes from *prov1* and *prov2*. Finally, on *d3*, the patient visits *prov1* twice and performs some procedures (*PX1* and *PX3*). Again, these two visits may or may not be related to each other. When submitting the insurance claims to *pt1*'s health insurance company, the providers aggregate service encounters at the day level, resulting in three medical claims (*clm1–clm3*) from *prov1*, one medical claim (*clm4*) from *prov2*, and one pharmacy claim (*clm5*) from *prov3*. Moreover, the claims from *prov1* and *prov2*, which originate from medical providers, are submitted through medical claims, whereas claims from *prov3* are submitted through pharmacy claims. For each claim, the provider can bill the insurance company for a certain amount (the *billed* amount), but the insurance company may pay all or part of that amount (the *paid* amount). Figure 1.b. shows the content of these five daily insurance claims. Since neither the visit identifiers nor physician notes are obtainable from daily claims, the connections among the visits, and those among the associated codes (medication, procedure, and diagnosis), are unknown. Also, neither the order of same-day visits nor the order of medical events within each visit are extractable from AC data.

Such lost relationships and unordered medical events become more problematic when there is a large number of single-day visits to multiple providers per day, which is most often the case for HN patients. This results in complex, heterogeneous patient profiles. In this study, we aim to reduce this heterogeneity through an effective deep learning representation learning architecture for AC data.

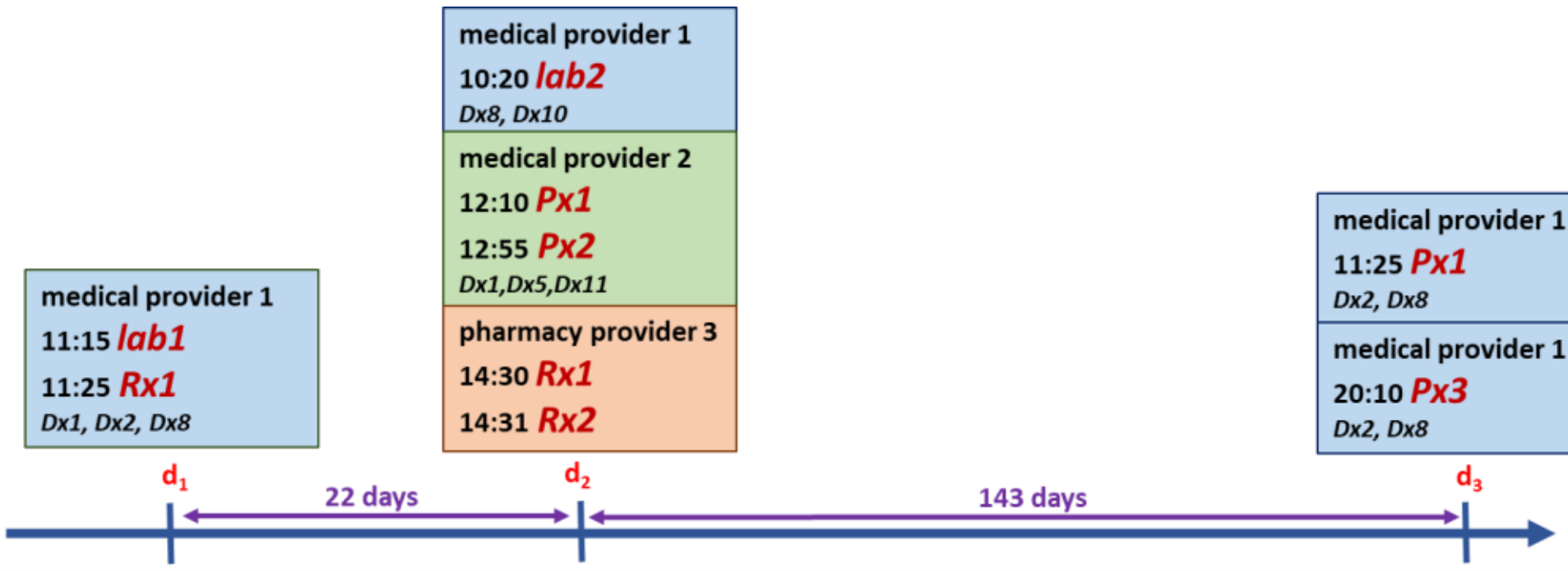

| Patient id | Visit id | Visit date | Visit admission time | Chart event id | Chart event time | Chart event type | Chart event result | Provider id |
|---|---|---|---|---|---|---|---|---|
| pt1 | vt1 | d1 | 11:00 | che1 | 11:15 | lab1 | xxx | prov1 |
| pt1 | vt1 | d1 | 11:00 | che2 | 11:25 | Rx1 | xxx | prov1 |
| pt1 | vt2 | d2 | 10:00 | che3 | 10:20 | lab2 | xxx | prov1 |
| pt1 | vt3 | d3 | 11:00 | che4 | 11:25 | Px1 | xxx | prov1 |
| pt1 | vt4 | d3 | 20:00 | che5 | 20:10 | Px3 | xxx | prov1 |

| Patient id | Visit id | Visit date | Visit admission time | Chart event id | Chart event time | Chart event type | Chart event result | Provider id |
|---|---|---|---|---|---|---|---|---|
| pt1 | vt1 | d2 | 12:00 | che1 | 12:10 | Px1 | xxx | prov2 |
| pt1 | vt1 | d2 | 12:00 | che2 | 12:55 | Px2 | xxx | prov2 |

| Patient id | Visit id | Visit date | Visit admission time | Chart event id | Chart event time | Chart event type | Provider id |
|---|---|---|---|---|---|---|---|
| pt1 | vt1 | d2 | 14:15 | che1 | 14:30 | Rx1 | prov3 |
| pt1 | vt1 | d2 | 14:15 | che2 | 14:31 | Rx2 | prov3 |

Figure 1.a. An example of patient *pt1*'s service visit history.

| Patient id | Claim id | Provider id | Billing date | Service date | Dx code 1 | Dx code 2 | Dx code 3 | ... | Dx code 10 | Px code | Amount |
|---|---|---|---|---|---|---|---|---|---|---|---|
| pt1 | clm1 | prov1 | xxx | d1 | Dx1 | Dx2 | Dx8 | ... | - | Px4 (lab1) | xxx |
| pt1 | clm1 | prov1 | xxx | d1 | Dx1 | Dx2 | Dx8 | ... | - | Px5 (Rx1) | xxx |
| pt1 | clm2 | prov1 | xxx | d2 | Dx8 | Dx10 | - | ... | - | Px6 (lab2) | xxx |
| pt1 | clm3 | prov1 | xxx | d3 | Dx2 | Dx8 | - | ... | | Px1 | xxx |
| pt1 | clm3 | prov1 | xxx | d3 | Dx2 | Dx8 | - | ... | - | Px3 | xxx |
| pt1 | clm4 | prov2 | xxx | d2 | Dx1 | Dx5 | - | ... | | Px1 | xxx |
| pt1 | clm4 | prov2 | xxx | d2 | Dx11 | Dx5 | - | ... | - | Px2 | xxx |

| Patient id | Claim id | Provider id | Billing date | Service date | Rx code | Amount |
|---|---|---|---|---|---|---|
| pt1 | clm5 | prov3 | xxx | d2 | Rx1 | xxx |
| pt1 | clm5 | prov3 | xxx | d2 | Rx2 | xxx |

Figure 1.b. An example of patient *pt1*'s heterogeneous AC profile. Note that each lab test or medication recorded by a provider has its own procedure or Px code, primarily for billing purposes.

**2.1.2. High-need patients.** Around 5% of the U.S. population accounts for approximately 50% of the nation's healthcare spending (Bélanger et al. 2019). As a result, much effort has been spent identifying patients with associated higher costs or needs in order to target them for cost-management programs and cost-reduction interventions (Anderson et al. 2015). To achieve this, the first step is to quantify the severity of patient needs according to numerical index measures. Researchers in (Joynt et al. 2017) grouped patients into six severity categories of need during a three-part series of workshops convened by the National Academy of Medicine. In ascending order of associated cost, there are six severity categories: (1) relatively healthy, (2) simple chronic, (3) minor complex chronic, (4) major complex chronic, (5) frail elderly, and (6) disabled. Table 1 describes these categories and Table A1 of Appendix A illustrates the included list of non-complex and complex chronic conditions, as well as frailty indicators. The precise method of assigning a patient to one of these six groups can be found in the original reference (Joynt et al. 2017).

In this study, HN patients refer to individuals belonging to severity categories (4), (5), and (6) in Table 1. It is important to note that extracting HN profiles necessitates extensive domain knowledge in order to identify the required severity groups using hundreds of diagnosis and procedure codes. Also, there are other methods of identifying HN patients based on various factors, such as their number of emergency room visits or unplanned hospital admissions (Heins et al. 2020), socio-behavioral or mental health conditions (Ford et al. 2017), and medical expenditures (Yang et al. 2018), which are not in the context of this study.

Table 1. Need severity categories extracted from (Joynt et al. 2017).

| **Relatively healthy (1)** | **Simple chronic (2)** | **Minor complex chronic (3)** | **Major complex chronic (4)** | **Frail elderly (5)** | **Disabled (6)** |
|---|---|---|---|---|---|
| 0 CCC & 0 NCC | 0 CCC & < 6 NCC | 1 or 2 CCC & < 6 NCC | ≥3 CCC & ≥ 6 NCC | ≥2 frail indicators | ESRD or Disabled |

CCC: Complex Chronic Condition – Non-complex Chronic Condition: NCC – ESRD: End-stage renal disease

Large amounts of the literature have focused on employing machine learning models to help HN patients for various applications, such as predicting their utilization of in-hospital resources (Yu et al. 2021); predicting their number of hospitalizations (Blumenthal et al. 2017, Al Snih et al. 2006), doctor visits (Al Snih et al. 2006), or emergency visits (Blumenthal et al. 2017); estimating their hospital readmission risk in the future (Billings et al. 2006); and clustering HN patients into subgroups who may benefit from particular care management interventions (Grant et al. 2020, Nuti et al. 2019) in effort to improve care coordination across settings (Yan et al. 2019).

While one of the primary objectives of RAMs is to reduce the favorable selection of healthier patients and avoid the disenrollment of HN patients, research has shown that, despite some success (McWilliams et al. 2012), these models have not been able to completely accomplish their goals (Jason Brown et al. 2014). Particularly, past studies have found that RAMs cannot accurately predict the costs of HN patients, leading CMS to underpay for these patients' health plans (Frogner et al. 2011, MaCurdy and Bhattacharya 2017,

Schillo et al. 2016) and incentivizing them to cherry-pick healthier enrollees by gaming the risk-adjustment scheme (Hellander, Himmelstein, et al. 2013). Moreover, healthcare providers with more HN populations have been found to be less likely to even switch into capitation-based payment models, since the associated higher levels of effort have not been financially rewarded (Rudoler et al. 2015).

Despite the variety of machine learning models that researchers have explored for RAMs, to the best of our knowledge, none of them has investigated their effectiveness for various levels of patients' need severity. We believe that the efficacy of machine learning models for HN patients with complex heterogeneous profiles is different from that of healthier patients, and thus the models' architectures need to be evaluated accordingly to consider the unique nature of HN profiles.

In this study we aimed at developing cost prediction models that are more accurate and equitable across different patient populations—especially HN patients—in order to address disparities in care. Therefore, the primary post-analysis evaluation design for this study focuses on investigating the performance of the proposed channel-wise method for patients with varying severity of need.

**2.1.3 Cost prediction models.** While predicting patient costs has always been important for various healthcare stakeholders, the recent impetus for government and commercial insurance providers to focus on RAMs has underscored its vitality (Wrathall and Belnap 2017). Nevertheless, as mentioned before, the performance of these models has not been wholly satisfactory, since the parametric ordinary least squares (OLS) models, which were initially used by RAMs, are limited in terms of extracting meaningful relationships among variables to make accurate predictions (Shrestha et al. 2018). Consequently, an emerging stream of literature has begun to explore the potential applications of machine learning methods that predict the healthcare costs of patients for risk-adjustment purposes (Kan et al. 2019).

Recent literature reviews on healthcare cost prediction (Morid et al. 2017) have revealed that prior research can fall into one of three motivational categories. First, studies have analyzed the importance of specific input variables on cost, rather than prioritizing the improvement of prediction performance itself (Eigner et al. 2019, Quercioli et al. 2018). Exploring the effect of an individual's neighborhood, including quality and location, using random forests (Mohnen et al. 2020) or examining a patient's chronic condition using regression trees (König et al. 2013) are examples of this type of research category. Second, research has directly improved cost prediction using only clinical variables of AC data, such as diagnosis codes (Duncan et al. 2016), procedure codes (Sushmita et al. 2015), medication codes (Jödicke et al. 2019), variety of medical codes (Yang et al. 2018), or comorbidity scores (Kuo et al. 2011). Third, the extant literature has predicted medical costs with both clinical variables (e.g., diagnosis codes, procedure codes, medication codes) and financial features, such as members' monthly costs using AC data (Kartchner et al. 2017). The proposed method in this current work can be classified under this third category of cost prediction literature. Consequently, this type of study has been reviewed in more detail throughout the rest of this section.

The most prevalent approach to applying traditional machine learning models to cost prediction involves feeding these models with aggregated patient annual data captured as static features, such as the total medical cost in the past or the total number of diagnosis codes (Morid et al. 2019). We have investigated a variety of traditional models—including linear regression (LR) (Kuo et al. 2011), gradient boosting decision tree (GBDT) (Duncan et al. 2016, Jödicke et al. 2019), CART (Bertsimas et al. 2008), random forest (RF) (Sushmita et al. 2015, Vimont et al. 2022), and the multi-layer perceptron (MLP) (Osawa et al. 2020, Rose 2016). (Morid et al. 2019) detected temporal change points in multi-variate time series and added change-point-related features to improve cost prediction performance. Recently, with the advent of deep learning models and their strength in capturing a variety of temporal patterns, a few studies have started employing them for cost prediction purposes. (Zeng et al. 2021) proposed a recurrent neural network (RNN) architecture, where clinical features are extracted as monthly temporal features, while financial features, such as cost, are extracted as historical static features. (Morid et al. 2020) proposed a convolutional neural network (CNN) architecture, in which each patient profile is considered as an image and then a 1D convolution model is trained on that input type. After conducting several sets of benchmarking experiments, these studies found that the representational learning power of deep architectures can outperform traditional machine learning models.

Despite their advantage, the current AC cost prediction deep learning models in the literature are not optimal for handling heterogeneous AC data, particularly AC data from HN patients. We discuss the deficiencies of these architectures, along with other AC deep learning architectures proposed for non-cost outcomes, in Section 2.2.2. Also, various cost prediction studies using deep learning and traditional machine learning models are benchmarked against the proposed architecture in this study. More specifically, since our focus is on cost prediction models for RAM, we selected the experimental benchmarks based on the applicability of their models for RAM.

## 2.2. Representation learning of patients' temporal data

Over the past decade, a variety of deep learning investigations have been conducted on healthcare predictive analytics with both supervised and unsupervised architectures. In order to distinguish architectures and their ultimate learning task, (Shickel et al. 2018) categorized these studies into five folds: information extraction, representation learning, outcome prediction, computational phenotyping, and clinical data de-identification. Information extraction focuses on extracting structural information from clinical notes. Representation learning aims to project discrete codes into continuous vector spaces for other downstream tasks. Outcome prediction concentrates on optimizing the various modules of a deep learning architecture to increase prediction performance for patient outcomes. Computational phenotyping seeks to derive data-driven descriptions of illnesses by revisiting and refining the broad boundaries of illness and diagnosis. Clinical data de-identification attempts to anonymize clinical notes that typically include explicit personal health information in order to facilitate their public release as useful clinical datasets. The focus of this study is on outcome prediction. It should be mentioned that certain representation learning studies do not stop at the patient representation step and are eventually optimized for a healthcare outcome prediction task. This can cause overlap in the above nomenclature—but to minimal effect.

Patient representations employed for deep time series prediction in healthcare can be classified broadly into two categories: *sequence representation* and *temporal-matrix representation* (Morid et al. 2022). In the former approach, which is more prevalent for the data collected on the healthcare providers' side (i.e., electronic health records, or EHR, data), each patient is represented as a sequence of ordered medical events. For medical-code sequences, each event exists as an ordered set of categorical medical codes (diagnosis, procedure, or medication). Since the lists of medical codes are generally quite long, various embedding techniques are commonly used to shorten them. Additionally, since medical events occur at different time intervals, the most common approach for handling the event irregularity is to concatenate the time interval between adjacent medical codes with the extracted embedding vectors. This approach has shown promising results with EHR data for predicting heart failure (Choi et al. 2017), vascular disease (Park et al. 2018), hospital mortality (Rajkomar et al. 2018), and hospital readmission (Rajkomar et al. 2018). However, to the best of our knowledge, this approach has not yet been evaluated for AC data applications.

Sequence embedding representations rely on two commonly used embedding methods. The first implements an embedding layer, trained with other network layers, to learn an embedded vector for a medical event or code (Choi, Bahadori, Schuetz, et al. 2016). This method is also called trainable embedding (Choi, Bahadori, Schuetz, et al. 2016). Otherwise, researchers typically invoke pre-trained embedding techniques related to Word2Vec, such as Skip-Gram (Choi, Bahadori, Schuetz, et al. 2016), for learning the embedded vectors for each medical code. Several studies have shown that using pre-trained models for medical information embedding outperforms learning from naïve embedding layers (Choi et al. 2017, Choi, Bahadori, Schuetz, et al. 2016, Maragatham and Devi 2019). Also, previous studies have shown that the embedding vectors of pre-trained models are transferable from one EHR to other data sources without losing performance (Choi, Bahadori, Schuetz, et al. 2016).

For the temporal-matrix representation category, each patient is represented as a longitudinal matrix, where rows correspond to regular time intervals and columns correspond to the variety of numerical medical measurements taking place during those time intervals. Resultingly, a single cell in a patient temporal matrix provides the coarse-grain aggregated value of a patient's numerical medical measurement during a particular time interval. The advantage of this approach is that, since patient records are divided into fixed-time windows, the event irregularity issue does not exist. However, the disadvantage is that not all patients have records for all time intervals, leading to missing value challenges. The most common approach for alleviating this challenge is to use masking vectors for learning data missingness (Y. W. Lin et al. 2019, Lipton et al. 2016, Tomašev et al. 2019). Also, using separate masking vectors for each input feature, in order to learn the missing patterns separately, has been effective as well (Harutyunyan et al. 2019).

Applying either sequence representation or temporal-matrix representation for payer-collected data (i.e., AC data) leads to an unignorable heterogeneity challenge. Unlike healthcare provider data (i.e., EHR data) where patients' profiles contain visit IDs as well as time-stamped medical events within each visit that are all related to each other, heterogeneous medical codes appearing within a day in AC data do not have any time stamp (i.e., order) and may be completely unrelated. As a stylized example—without stretching credulity—suppose a patient record for the day shows diabetes diagnosis codes, along with procedure codes for dental CT scans and pharmacy codes for anti-allergy medications. These disparate entries are obviously related to medical events of different origins. In actuality, the patient could have had a dental visit a few days prior but may have wanted to order the CT scan for the same day as the diabetes appointment, while concomitantly making the effort to refill the anti-allergy medication on the way home from radiology. This heterogeneity challenge becomes more intensive for HN patients with a significant number of multi-provider, same-day visits. Moreover, previous studies have aggregated the daily granular AC data to grains like weekly granular (Fenglong Ma et al. 2017) or, particularly for cost prediction, monthly granular (Morid et al. 2020, Zeng et al. 2021), which can further intensify the heterogeneity issues, especially for HN patients with complex patient profiles. Consequently, one research gap that this study addresses involves finding an appropriate deep learning architecture that can better learn patient representation from daily AC data to reduce these types of heterogeneity effects.

## 2.3. Healthcare predictive analytics in information systems

The machine learning models, techniques, and systems developed to address healthcare needs have become a recent focus of IS researchers (Y. K. Lin et al. 2019, Liu et al. 2020, Zhu et al. 2021), particularly due to the increased interest in big data and predictive analytics (Chen et al. 2012, Shmueli and Koppius 2011). Many research studies are motivated by the desire to leverage traditional machine learning or deep learning methods on rich data sources that were previously unavailable or underutilized (Lin et al. 2017). These data sources are usually collected from the digital traces of patients stored as longitudinal medical records (Hedman et al. 2013). Examples of this research stream include (1) deep learning methods for recognizing the activities of daily living in senior care facilities (Zhu et al. 2021) or detecting their fall incidents (Yu et al. 2023), (2) Bayesian multitask learning for the risk profiling of chronic patients (Lin et al. 2017), and (3) deep neural networks for examining user engagement with encoded medical information in YouTube videos (Liu et al. 2020). While this study is a continuation of such healthcare studies in the way it benefits from advanced machine learning methods to enhance representation learning and temporal pattern extraction, there are some key differences. Particularly, in some of the prior studies, the data was collected from a single source, such as EHR (Lin et al. 2017), or belonged to individual subjects, such as senior citizens (Zhu et al. 2021), with unambiguous connections among medical events and accurate time stamps for temporal learning. In contrast, this study's data comes from multi-source AC data with unknown links between daily submitted claims and daily granular time stamps, leading to the heterogeneity challenge. This study aims to investigate how alleviating this heterogeneity with proper representation learning affects the performance of the underlying prediction task, leading to enhanced decision-making power.

Furthermore, the effective use of healthcare resources and the management of patient costs have been additional focal points for IS researchers. Prototypical works include (1) cost-effective clinical decision-making for removing patients from mechanical ventilation (Fang et al. 2021), (2) predicting and controlling future patient readmission (Bardhan et al. 2014, Ben-Assuli and Padman 2020), and (3) reducing patient pharmacy costs by providing associated cost information to physicians (Bouayad et al. 2020). While this study continues the same goal of efficiently using available resources, the current work focuses more on decision support for payers and health plans in the healthcare ecosystem.

Finally, a related IS research stream studies the socio-technical context of user digital traces (Berente et al. 2019, Xie et al. 2022). This enables researchers to analyze the technical aspects of predictive analytics models based on the social dimensions of user behavior (Abbasi et al. 2019, Li et al. 2020, Liu et al. 2020, Zhang and Ram 2020). This study adds to this stream with the technical methodology design of a deep learning architecture that reduces AC data heterogeneity. It also adds humanistic-impact evaluation design to investigate the effect of the proposed method on heterogeneous patients. Compared to the IS literature that incorporates patients' digital traces (Pentland et al. 2020), this study focuses on patient experience via multi-source AC data, rather than single-source data such as EHR (see Section 2.2). Furthermore, in contrast to such studies that have concentrated on process mining tasks, this study specifically centers on a prediction task. Nevertheless, the digital trace patterns extracted from patients' profiles can also provide valuable insights and expand the scope of research on process mining tasks.

# 3. Guiding Socio-technical Considerations and Research Gaps

Consistent with the methods of prior studies (Bertsimas et al. 2008, Morid et al. 2017), solving the healthcare cost prediction problem involves providing an annual individual cost prediction model that can be applied to the insured population. For each individual, it estimates the amount of total healthcare expenses incurred by the individual over the next year using the medical codes, associated expenses, and day stamps available in each of the ACs submitted for the individual in the prior year. This study includes the expenses associated with the entire variety of medical codes—diagnostic and procedure as well as medication codes. Because the poor performance of cost prediction models on HN patients originating from their heterogeneous medical journeys was the original motivation for the proposed technical design, the social and technical aspects of this study have been inextricably linked since its inception. Table 2 summarizes the guiding socio-technical desiderata for the design and evaluation of the proposed cost prediction architecture and the corresponding research gaps we will address to accomplish these desiderata.

Critical to principled capitation decision-making, accurate patient cost predictions—the first guiding social desideratum—must be achieved with multi-source insurers' AC time series data. The social and contextual factors in patients' complex medical journeys have resulted in a variety of data heterogeneity manifestations, including different semantics of codes from multiple coding systems, unknown inter-visit relationships, and orderless codes within a single claim. We employ a multi-pronged approach to reducing the various data heterogeneity issues that the current literature on cost prediction and other health outcome predictions has not satisfactorily overcome. To address the challenge of learning from different kinds of code and their associated costs in claims for the same patient, the proposed prediction architecture first learns signals in a homogeneous code or cost channel before fusing the resulting representations from all channels. Section 4 describes how this study uniquely extends this channel-wise learning design to effectively fit code types and the associated costs to multiple channels of sequential learning so as to initially focus on temporal signals and the differential importance of intra-channel data heterogeneity in representation learning with homogeneous data (i.e., one code or cost type) only.

Table 2. Guiding socio-contextual and technical desiderata for cost prediction design and evaluation.

| **Social desiderata** | **Design-evaluation desiderata** | **Relevant research gaps** |
|---|---|---|
| FFV implementations rely on accurate cost predictions using AC data available to payers. | Design and implement effective cost predictions to address heterogeneous data challenges resulting from AC data limitations of unknown inter-visit relationships and orderless, multi-channel codes in claims. | The accuracy of extant cost prediction models needs improvement.<br><br>Deep learning models designed for other health outcome predictions based on same-source data are not suited for cost predictions due to AC data limitations. |
| Reduce payment discrepancy, especially for HN patients who are disproportionally affected by poor cost prediction accuracy. | The reduction in payment discrepancy increases with the complexity of patient needs.<br><br>Strive for high coherence for same need level patients' representations and high distinctness otherwise. | Past studies on HN patients did not adequately design cost-prediction models to benefit HN patients.<br><br>Evaluations of extant cost-prediction models did not assess the reduction of payment disparity and patient representation coherence for HN patients. |
| High volume of claims and diversity of providers and medical codes further delineate high data heterogeneity of HN patients. | Appropriately measure data heterogeneity in multi-channel time series of medical codes. | Current entropy indexes are not designed to measure the heterogeneity of multi-channel categorical time series data. |
| An efficient patient segmentation method that is well aligned with patient need levels affords flexible, scalable analysis of differential cost-prediction accuracy and its impact on payment discrepancy reduction over more varieties of patient categorization. | In light of the alignment of medical journey complexity and data heterogeneity, utilize an appropriate multi-channel entropy index to segment patients and evaluate cost prediction accuracy and payment discrepancy reduction over approximated segments of patient need levels. | Past literature has not probed the potential of approximating patient need segments with a patient AC data heterogeneity index. |

The second social desideratum focuses on reducing payment discrepancies when the actual amount of total payments that a patient's insurance plan makes is higher or lower than the predicted amount (i.e., the capitation amount). By effectively reducing data heterogeneity, the proposed architecture is designed not only to improve the accuracy and downstream impact for all patients, but also to foster more significant improvements for HN patients. To confirm the alignment between the varying complexities of patient journeys and patient representation learning, we visualize the extent of patient representation coherence, or similarity, for patients of the same need levels.

To probe how patient characteristics—such as the volume, frequency, and diversity of the code, claims, and providers of patient journeys—affect the reductions in prediction errors and payment discrepancies, the third technical desideratum involves quantitatively measuring data heterogeneity in multi-source time series of codes (or categorical data in other contexts). Because the most common heterogeneity measurement—the entropy index—is not readily applicable, we define a multi-channel heterogeneity index for delineating patients' data heterogeneity levels that can closely approximate their health severity levels in Section 6.

The last social desideratum addresses two challenges of performing analyses for patients at different severity or need levels: (1) the extant codifications of patient severity categories critically depend on domain knowledge and require additional access to lists of conditions in order to codify patients by these categories and (2) the complex process of developing such codifications results in bottlenecks in analyzing the differential impact of decisions on heterogeneous patients. Section 6 and Appendix E further demonstrate that, when the complexities of patients' medical journeys to address their health needs are closely aligned with AC data heterogeneity, evaluations of cost prediction models can utilize the multi-channel data heterogeneity index to help stratify heterogeneous patients with greater flexibility and scalability. Overall, this study fulfills a programmatic need by providing evaluation insights into the role and benefit of an adequate multi-source data heterogeneity abstraction.

# 4. Model Development

In this research, we first assert that a channel-wise architecture is needed for efficient representation learning, in which heterogeneous medical codes occurring within a day are separated by the type of medical code (diagnosis, procedure, or prescription)—as well as the type of cost coming from each type of provider—in order to reduce the heterogeneity of temporal patterns over different types of code. This study's sequence-representation-based AC channel-wise architecture, which alleviates heterogeneity, is motivated by temporal-matrix representation-based EHR channel-wise architecture that addresses missingness (Harutyunyan et al. 2019) (Section 2.2). Second, to avoid generating heterogeneity through coarse-grain aggregation, we leverage the natural time series at the day level of AC data, the finest available granularity, for accurate prediction. This can be accomplished with a pre-trained embedding of daily medical events, a choice motivated by EHR studies demonstrating that pre-trained medical-code embeddings are more accurate than embeddings achieved with trainable embedding layers, which also provide knowledge transfers between different health information systems (Section 2.2). As opposed to the Word2Vec pre-trained embedding, which relies on ordered visits and ordered medical codes unavailable in AC data, this study adopts a Doc2Vec pre-trained embedding, which does not rely on the order of medical codes within daily claim events of AC data. Third, we employ an attention layer to combine the temporal outputs of each channel and weigh the extracted daily event representations according to their importance for the final prediction. Aligned with the recent trend of deep learning architectures in healthcare (Morid et al. 2022), the attention weights of daily events at the patient level can emphasize different parts of patient journeys and enhance prediction performance for heterogeneous patient profiles.

A general framework for the proposed architecture is shown in Figure 2. The proposed deep time series prediction architecture consists of three components: (I) channel-wise representation using Doc2Vec embedding, (II) temporal pattern extraction using RNN, and (III) the channels' output combination for the final prediction using attention. This section first elaborates on each of these components in detail and then illustrates the overall channel-wise architecture.

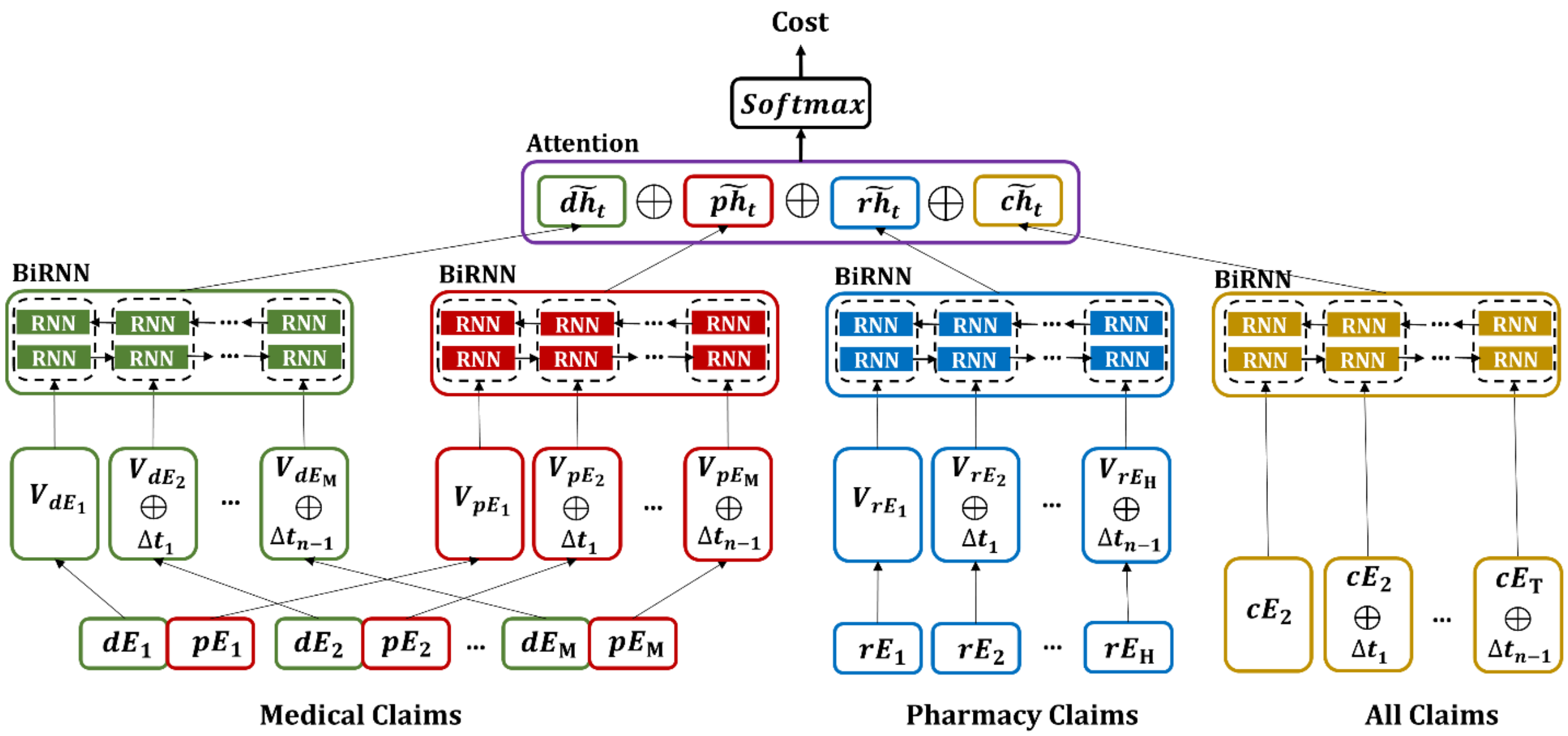

Figure 2. The proposed AC deep learning architecture with channel-wise representation learning.

## 4.1. Medical-event embedding

As mentioned, in order to use a pre-trained embedding method that aligns with the nature of AC data, where the order of visits and medical codes is missing, we leveraged the Doc2Vec technique, particularly the Paragraph Vector for Distributed Bag of Words (PV-DBOW) model (Le and Mikolov 2014), to generate embedding vectors for medical and pharmacy events associated with AC data. In the natural language processing literature, a PV-DBOW model receives each document as an ordered collection of paragraphs, in which each paragraph is an unordered set of words, and then learns to predict the embedding of each paragraph. We adopt this method for AC data by considering each patient as an ordered collection of daily medical events, in which each medical event is an unordered set of medical codes. Once embedded, a medical event, $E_i$, has a vector representation denoted by $V_{E_i} \in \mathbb{R}^m$, where $m$ is the embedding dimension. To the best of our knowledge, this is the first study to use pre-trained medical-event embedding—as opposed to trainable medical-code embedding—for AC data.

## 4.2. Bidirectional recurrent neural networks

Bidirectional recurrent neural networks (BiRNNs) are the most successful state-of-the-art RNN models for patient time series data, outperforming vanilla RNNs in several studies (Barbieri et al. 2020,

Fenglong Ma et al. 2017, Wei Guo et al. 2019, Harutyunyan et al. 2019, Sun et al. 2019). A BiRNN consists of a forward and backward RNN, where the forward RNN reads the input vector of medical events from $V_{E_1}$ to $V_{E_T}$ and calculates a sequence of forward hidden states $\overrightarrow{h_1}, \ldots, \overrightarrow{h_T}$ ($\overrightarrow{h_\iota} \in \mathbb{R}^p$), while the backward RNN reads the vector of medical events in reverse order, creating a sequence of backward hidden states $\overleftarrow{h_1}, \ldots, \overleftarrow{h_T}$ ($\overleftarrow{h_\iota} \in \mathbb{R}^p$). Here, $p$ is the dimensionality of hidden states. By simply concatenating the forward and backward hidden state vectors, the final hidden state vector representation becomes $h_i = [\overrightarrow{h_\iota}; \overleftarrow{h_\iota}]^\top$ $h_i \in \mathbb{R}^{2p}$.

GRU networks differ from vanilla RNN networks in that they have an internal constitution of reset and update gates, which allow the network to discard past information that becomes less relevant, update new information that becomes important, and retain pertinent information from one timestep to another within the network's sequential temporal operations. The GRU architecture successfully allows the network to avoid the vanishing or exploding gradient problem inherent in traditional RNNs, and therefore it retains their learned information over many time steps (Yin et al. 2019). In this paper, we adopted Bidirectional GRU (BiGRU) as the specific deep learning model used separately in each channel and also in the final aggregate.

## 4.3. Attention mechanism

There exists a recent trend of adding attention layers to RNN models in healthcare deep time series prediction literature (Morid et al. 2022), since attention captures patterns between medical events and also emphasizes different parts of patient journeys to enhance prediction performance for heterogeneous patient profiles. In this study, we adopted a concatenation-based attention (Bahdanau et al. 2015) to extract the importance of each medical event as follows:

$$\alpha_{ti} = v_\alpha^\top \tanh(W_\alpha[h_t; h_i]) \tag{1}$$

$$\alpha_t = \frac{e^{\alpha_{ti}}}{\sum_{i=1}^{t-1} e^{\alpha_{ti}}} \tag{2}$$

$$c_t = \sum_{i=1}^{t-1} \alpha_{ti} h_i \quad (3)$$

Here, $W_\alpha \in \mathbb{R}^{q \times 4p}$ and $v_\alpha \in \mathbb{R}^q$ are network parameters to be learned, where $q$ is the latent dimensionality. Also, $\alpha_{ti}$ is the attention weight vector of hidden state $h_i$, $\alpha_t$ is the attention weight vector of all hidden states representing medical events, and $c_t$ is the context vector for time *t*. The resulting attentional hidden state captures context and hidden state information via a concatenation operation from the context vector, $c_t$, and the current hidden state, $h_t$:

$$\widetilde{h_t} = \tanh(W_c[c_t; h_t]) \quad (4)$$

where $W_c \in \mathbb{R}^{r \times 4p}$ is the weight matrix, $\widetilde{h_t}$ is the final attention vector, and *r* is the dimensionality of $\widetilde{h_t}$. It should be mentioned that we also tried location-based attention mechanisms, but they were not as effective as the above concatenation-based attention schema, since location-based mechanisms do not consider cross-patterns between medical events.

## 4.4. Channel-wise learning

Assuming a given total of *N* patients, the *n-th* patient has $M^{(n)}$ medical events (days), each of which includes one or more medical claims, and $H^{(n)}$ pharmacy events (days), each of which includes one or more pharmacy claims. The coded representation of the $n$-th patient can be constructed by medical and pharmacy codes using a sequence of diagnosis events (days), $dE_1, dE_2, \ldots, dE_{M^{(n)}}$, a sequence of procedure events (days), $pE_1, pE_2, \ldots, pE_{M^{(n)}}$, and a sequence of medication events (days), $rE_1, rE_2, \ldots, rE_{H^{(n)}}$. Recall that each event is a day in which one or more diagnosis, procedure, and/or medication codes may appear. In addition to combining medical and pharmacy costs together for the purposes of assessing patient cost, a patient can also be represented further by a sequence of cost events (days), $cE_1, cE_2, \ldots, cE_{T^{(n)}}$, $E_i \in \mathbb{R}$ , where $T^{(n)}$ is the total number of days in which patient $n$ has made one or more claims, be they related to procedures or medications. In other words, each cost event, $cE_t$, is a vector containing medical costs concatenated with pharmacy costs in the form of a dollar amount recorded on day $t$. For simplicity, we describe the proposed architecture for a single patient below, and we thus drop superscript *(n)* in the rest of the paper. All patient data is exposed to the same pipeline of mathematical operations and transformations.

For the proposed channel-wise learning architecture in this paper, each individual categorical feature—including diagnosis codes ($dE_M$), procedure codes ($pE_M$), and medication codes ($rE_H$)—is first embedded within an event using the aforementioned PV-DBOW method to generate the corresponding $V_{dE}$ , $V_{pE}$ , and $V_{rE}$ embedding vectors. Next, the time interval between a claim event (i.e., diagnosis, procedure, or medication event), which is the collection of claims received on that event (day), and its previous claim event is concatenated to the trio of associated embedding vectors: [$V_{dE}$; Δt], [$V_{pE}$; Δt], and [$V_{rE}$; Δt]. Note that time intervals coming from pharmacy claims will differ with those from medical events, as diagnoses and procedures are connected in a many-to-one fashion in AC data, while pharmacy claims are uncoupled from the claims of the other two categories. Furthermore, the time interval between claim events is added to the embedded daily-cost-feature vectors [$cE_T$; Δt]. These four input vectors are then passed to a BiRNN layer and enhanced with concatenation-based attention, which extracts temporal patterns from each individual categorical-claim feature type, as well as the cost features, in order to generate the preliminary attention vectors, denoted respectively as $\widetilde{dh_t}$, $\widetilde{ph_t}$, $\widetilde{rh_t}$, and $\widetilde{ch_t}$. Finally, the four attention vectors are concatenated in an ordered manner, [$\widetilde{dh_t}$; $\widetilde{ph_t}$; $\widetilde{rh_t}$; $\widetilde{ch_t}$], to build the final attentional vector, $\widetilde{h_t}$, which is then given to a dense linear neural-network layer to predict the total patient cost for the following annum:

$$Cost_t = W_s\widetilde{h_t} + b_s \quad (5)$$

where $W_s$ and $b_s$ are network parameters to be learned.

# 5. Data, Experiments, and Results

## 5.1. Data and experimental setup

Our primary dataset consisted of 7.1 million medical claims and 1.8 million pharmacy claims from approximately 111,000 Medicare patients who resided in Utah from January 2015 to December 2018 and had at least two claims per year. This AC data consists of diagnosis, procedure, and medication codes as well as cost information—including allowed, paid, and billed amounts.

As in previous studies on cost prediction (Bertsimas et al. 2008, Morid et al. 2019), we used one prior year of patient data, the designated observation period, to predict patient cost in the following year, the result period. Specifically, the data in the years 2015, 2016, and 2017 was used to predict costs in the years 2016, 2017, and 2018, respectively. Further aligning with related papers in the extant cost prediction literature (Bertsimas et al. 2008, Duncan et al. 2016, Frees et al. 2013), the transaction cost of a particular medical service or product, such as a prescription or lab test, was measured in terms of the dollar amount paid for that medical service or product. The target variable is then an aggregation of these claims costs, amounting to the total annual liability incurred during the result period.

For all experiments, we randomly selected 60% of the data for training, 20% for validation, and 20% for testing. We performed hyperparameter tuning on the validation set, and we used optimal values to report the performance on the testing set. To improve experimental robustness, we employed a repeated random sub-sampling for the validation process, where training/validation/testing splits were randomly shuffled twenty times each. Except for the monetary analysis, we measured prediction performance according to the mean absolute percentage error (MAPE):

$$MAPE = \frac{1}{N}\sum_{i=1}^{N}\left|\frac{Actual_i - Predicted_i}{Actual_i}\right| \quad (6)$$

Here, $N$ is the number of patients in the testing dataset, and $Actual_i$ and $Predicted_i$ are the actual and predicted annual cost values for each patient, respectively. We applied a Wilcoxon signed-rank test (Wilcoxon 1992) to test the statistical significance of the differences in MAPE among the various methods.

For the monetary analysis in Section 5.5, the mean absolute error (MAE) of predictions, total overpays, total underpays, and net pay were calculated as follows:

$$MAE = \frac{1}{N}\sum_{i=1}^{N}|Actual_i - Predicted_i| \quad (7)$$

$$Underpay = \sum_{i=1}^{N} Actual_i - Predicted_i\,, if\ Actual_i > Predicted_i \quad (8)$$

$$Overpay = \sum_{i=1}^{N} Predicted_i - Actual_i \text{ , } if \ Actual_i < Predicted_i \quad (9)$$

$$Netpay = Overpay + Underpay \quad (10)$$

## 5.2. Benchmark analysis

**5.2.1. Channel-wise learning cost prediction versus cost prediction methods in the literature.** Since this study focuses on predicting cost using clinical and non-clinical (financial) patient features, we extracted the first set of benchmarks from the cost prediction literature (Section 2.1.1). Particularly, we used two deep learning architectures—including CNN (Morid et al. 2020) and RNN (Zeng et al. 2021)—as well as traditional machine learning models—such as LR (Kuo et al. 2011), GBDT (Jödicke et al. 2019, Morid et al. 2019), CART (Bertsimas et al. 2008), RF (Sushmita et al. 2015), and MLP (Osawa et al. 2020)—as benchmarks. All models were implemented with the exact input features used in the cited studies.

Table 3 shows the performance comparison between the proposed channel-wise deep learning model versus the benchmark cost prediction models in the literature (Section 4.5.1). Aligned with previous deep learning studies, we observed that extracting clinical and non-clinical patterns using the representation learning power of deep learning models can outperform traditional machine learning (Morid et al. 2020, Zeng et al. 2021) ($p<0.01$).

Table 3. Total cost prediction for channel-wise learning versus cost prediction models in the literature.

| Reference | Model | Clinical Features | Financial Features | MAPE |
|---|---|---|---|---|
| (Kuo et al. 2011) | LR | Static | Static | 89.3 |
| (Sushmita et al. 2015) | RF | Static | Static | 86.0 |
| (Bertsimas et al. 2008) | CART | Static | Static | 84.8 |
| (Jödicke et al. 2019) | GBDT | Static | Static | 83.8 |
| (Osawa et al. 2020) | MLP | Static | Static | 83.7 |
| (Morid et al. 2019) | GBDT | Temporal | Static | 83.4 |
| (Zeng et al. 2021) | RNN | Temporal | Static | 78.9 |
| (Morid et al. 2020) | CNN | Temporal | Temporal | 62.3 |
| Channel-Wise | RNN | Temporal | Temporal | 46.8 |

Moreover, when comparing (Morid et al. 2020) against (Zeng et al. 2021), it was observed that cost should be captured as a temporal feature—as opposed to a static feature—in order to better leverage its historical patterns. The proposed channel-wise architecture was able to outperform (Morid et al. 2020) with a large margin of 15.5% MAPE. This result is rooted in three main differences between the two compared models. First, (Morid et al. 2020) uses a temporal-matrix representation, which involves heavy monthly data aggregation and corresponding information loss. Second, their method combines all types of medical codes for each time step in a one-dimensional kernel, thereby contributing to a heterogeneity increase that misleads the predictive model. Third, the CNN employed by (Morid et al. 2020) and the adopted RNN model of the current study extract different kinds of temporal patterns. The former focuses on local motif features within a fixed-length window size, while the latter captures long-term trends (Ma et al. 2018, Morid et al. 2022).

**5.2.2. Channel-wise learning cost prediction versus non-cost prediction methods in the literature.** Due to the fundamental differences between the nature of the data collected on the healthcare provider's side (i.e., EHR) and the healthcare payer's side (i.e., AC), machine learning methods that are successful on EHR data may not work well on AC data. For example, deep learning methods have been effective for patient readmission prediction using EHR data (Cheung and Dahl 2018, Y. W. Lin et al. 2019, Nguyen et al. 2017, Wang et al. 2017), but they do not provide promising results for a corresponding AC prediction task (Min et al. 2019). In this study, we benchmarked the proposed method against some EHR baselines in order to shed light on the limited applicability of EHR-based, deep time series prediction methods to AC data for the cost prediction task. Therefore, while the focus of this paper is on cost prediction using AC data, another set of benchmarks in this study include various state-of-the-art deep learning architectures proposed for AC or EHR data—designed for healthcare outcome predictions that are not cost-related. The benchmark studies selected for this purpose include (Choi, Bahadori, Schuetz, et al. 2016), also known as Doctor AI; (Choi, Bahadori, Kulas, et al. 2016), also known as RETAIN; (Fenglong Ma et al. 2017), also known as Dipole; (Choi, Schuetz, Stewart, et al. 2016); and (Rajkomar et al. 2018). Furthermore, in order to provide a complete picture of the various architectures, the deep learning architectures from Section 5.2.1 that were proposed for cost predictions (Morid et al. 2020, Zeng et al. 2021) are also included in the comparison.

Since the above benchmarks are not directly applicable to AC cost prediction, we implemented three changes in their architectures to accommodate this task. First, these studies mainly use categorical medical codes as input, so the efficient incorporation of numerical cost data with the embedding vectors of medical codes was a technical challenge. Because EHR studies—when adding numerical features, particularly time intervals (between patient events)—concatenate the numerical values to the end of the embedding vectors (Section 2.2.2), we assumed that treating costs in the same way would be a trivial approach to addressing the challenge and further strengthening the benchmarking experiments. Second, since EHR studies need ordered sets of visits as input, and because individual visits are not extractable from AC data, we considered each claim day as a single medical event to be treated as an input visit. Third, since EHR studies with pre-trained embedding need ordered sets of medical codes, whereas AC data does not include the timestamp of medical codes, we ordered them randomly within a claim day.

To ensure an objective comparison of all the various deep learning architectures, each benchmark is implemented with precisely two layers of BiGRU. Recent studies have shown the advantage of a double-layered architecture over a single-layer architecture (Choi, Bahadori, Schuetz, et al. 2016, Fenglong Ma et al. 2017).

Table 4 displays the results of various deep learning architectures versus the proposed channel-wise architecture. All benchmarks are outperformed by the proposed architecture ($p<0.01$). Specifically, this success can be attributed to several key factors: (I) using multi-channel versus single-channel RNN temporal-pattern extraction of medical codes, (II) using fine-grain daily data versus coarse-grain weekly or monthly data, (III) using pre-trained (Doc2Vec) versus trainable embedding, and (IV) capturing cost information with a separate channel versus simply adding event cost to the embedding vectors. These four components are further explored in Section 5.6 and Appendix C to better explain their respective contributions. Also, detailed descriptions of these studies, along with a literature review of AC and EHR deep learning models, can be found in Appendix G.

Table 4. Total cost prediction for channel-wise learning versus deep time series architectures in the extant literature.

| Arch | MedCodeT | MedVisT | TimeInt | Granularity | EmdSubj | EmdApp | EmbUnit | PatRep | UP | Model | Att | MAPE |
|---|---|---|---|---|---|---|---|---|---|---|---|---|
| (Rajkomar et al. 2018) | Y | N | Fixed window | Hour | Medical code | Trainable | Individual medical codes and lab values | SeqRep | Concatenated average of embedding vectors per hour | RNN | Y | 88.9 |
| (Choi, Schuetz, Stewart, et al. 2016) | Y | N | N | Medical code | Medical code | Word2Vec | Bag of codes | SeqRep | Sum of embedding vectors per visit | RNN | N | 87.7 |
| (Zeng et al. 2021) | Random | N | Fixed window | Month | Medical code | Word2Vec | Bag of codes | SeqRep | Sum of embedding vectors per month | RNN | N | 80.9 |
| (Choi, Bahadori, Schuetz, et al. 2016) | Y | N | Concatenated with embedding vectors | Medical code | Medical code | Word2Vec | Bag of codes | SeqRep | Embedding vector per medical code | RNN | N | 66.9 |
| (Morid et al. 2020) | N | N | Fixed window | Month | - | - | - | TmpMax | Count of medical codes per month | CNN | N | 62.3 |
| (Fenglong Ma et al. 2017) | N | N | Fixed window | Week | Weekly claims | Trainable | Bag of weekly claims | SeqRep | Embedding vector per week | RNN | Y | 60.5 |
| (Choi, Bahadori, Kulas, et al. 2016) | N | Y | N | Visit | Medical visits | Trainable | Bag of visits | SeqRep | Embedding vector per visit | RNN | Y | 59.4 |
| Channel-Wise | N | N | Concatenated with embedding vectors | Day | Daily claims | Doc2Vec | Separate bag of code types | SeqRep | Embedding vector per day per medical code type | Channel-wise RNN | Y | 46.8 |

**Arch:** Architecture. **MedCodeT**: Using the medical code timestamp. **MedVisT**: Using the medical visit timestamp. **TimeInt**: Incorporating the time interval between visits. **EmdSubj**: Embedding subject. **EmdApp**: Embedding approach. **EmbUnit**: Embedding unit. **PatRep**: Patient representation. **UP**: Unit of processing for the deep learning model at each time step. **Att**: Attention model. **SeqRep**: Sequence representation. **TmpMax**: Temporal-matrix representation.

## 5.3. High-need analysis

To evaluate the effectiveness of the proposed architecture in estimating costs for HN patients, all individuals in the test dataset were partitioned into the six severity categories introduced in (Joynt et al. 2017) (Section 2.1.2). The claims cost prediction performance was then calculated for each severity category separately and compared against baseline models.

As evidence of the heterogeneity of HN patient profiles, Table 5 shows the average number of visited providers per day, week, and month over HN categories, along with other descriptive statistics. Although, aligned with our premise, we can broadly conclude that HN patients with more heterogeneous profiles contain more same-day, same-week, or same-month visits, this still may not provide a precise picture of daily patient experience, and the heterogeneity may be even worse for three reasons. One, because providers cannot submit multiple claims for a single patient per day, multiple visits to the same provider on a single day might be counted as a single claim. Two, although multiple provider ID records on the same day might still be related, this becomes less likely for highly heterogeneous groups who average 7.84 daily visits. Three, a patient purchasing multiple medications on a single pharmacy visit may accumulate pending medications that were not picked up during past visits, possibly due to backorders or other factors. All such medications are counted as a sole provider ID record. More details on the sensitivity analysis over the granularity aspect of the proposed method can be found in Appendix C.

Table 5. Descriptive statistics of patients' medical and pharmacy claims over need severity categories.

| | **Granularity** | **Relatively healthy** | **Simple chronic** | **Minor complex chronic** | **Major complex chronic** | **Frail elderly** | **Disabled** |
|---|---|---|---|---|---|---|---|
| Population (%) | - | 11 | 19 | 26 | 18 | 17 | 9 |
| Average number of providers | Day | 1.75 | 2.31 | 3.09 | 4.69 | 6.14 | 7.84 |
| | Week | 1.78 | 2.39 | 3.24 | 5.16 | 6.89 | 8.96 |
| | Month | 1.89 | 2.58 | 3.58 | 5.94 | 8.54 | 11.78 |
| Average number of unique clinical codes | Month | 4.14 | 8.28 | 13.17 | 25.11 | 33.17 | 35.03 |
| Average number of claims | Month | 2.96 | 9.06 | 18.84 | 28.13 | 39.36 | 46.11 |
| Average number of claim events | Months | 0.84 | 1.22 | 2.64 | 3.45 | 3.84 | 4.28 |

Table 6 displays the results of total claims cost prediction over the six severity categories of need severity in (Joynt et al. 2017). Aligned with previous literature (Frogner et al. 2011, MaCurdy and Bhattacharya 2017), this table shows that predicting the future costs of patients with higher needs is increasingly more difficult ($p<0.01$). Nevertheless, the proposed channel-wise learning benefits over single RNN learning in a correspondingly increasing manner, from 2.4% to 27.2% over the six need severity categories. This validates our premise that reducing the AC data heterogeneity by separating representation learning and temporal pattern extraction for each medical code type can improve the prediction performance, an effect which is more significant for higher heterogeneous profiles.

Table 6. Total cost MAPE performance over need severity categories.

| Learning | Relatively healthy | Simple chronic | Minor complex chronic | Major complex chronic | Frail elderly | Disabled |
|---|---|---|---|---|---|---|
| Single channel | 31.3 | 37.4 | 49.9 | 70.4 | 93.2 | 116.9 |
| Channel-wise | 28.9 | 30.5 | 38.5 | 47.3 | 67.4 | 88.7 |
| **Improvement (Difference)** | **2.4** | **6.9** | **11.4** | **23.1** | **25.8** | **27.2** |

To shed light on the internal procedure of the proposed channel-wise architecture compared to the baseline single-channel architecture, we conducted four visualization experiments. First, to visually illustrate the representational learning power of the proposed approach, Figure 3 depicts the final patient representation vectors, color-coded for patient HN severity category, for channel-wise versus single-channel architecture in 2-D space with t-SNE after k-medoids clustering (Maaten and Hinton 2008). Cluster centers were initialized with one random HN patient from each severity category. As the figure shows, the proposed architecture provides less complex input vectors that are more distinguished for heterogeneous patient profiles in the final prediction layer.

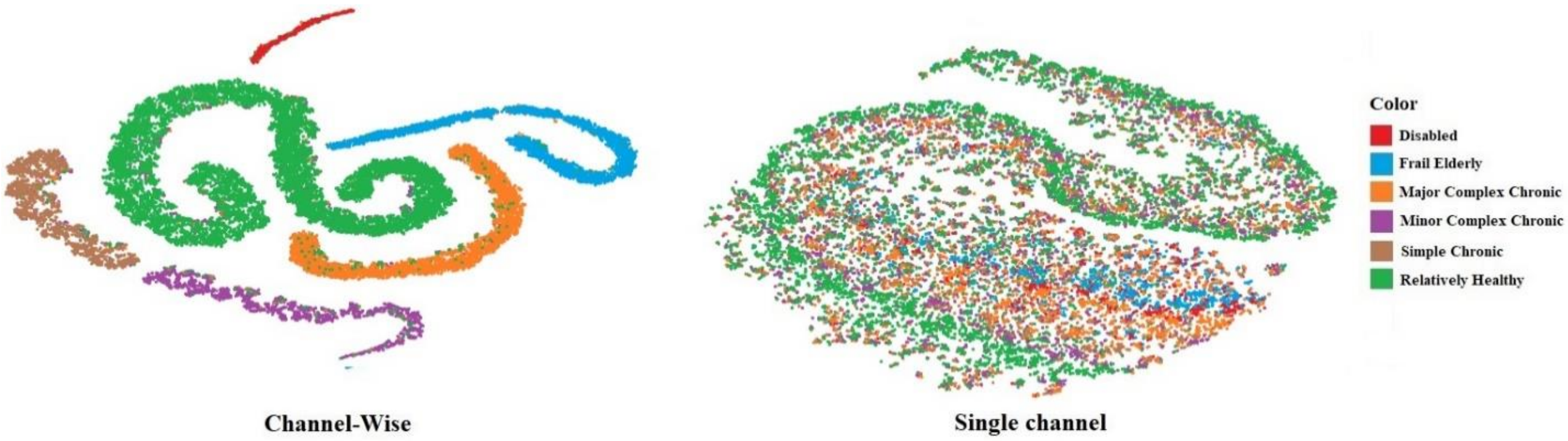

Figure 3. Representation vector of channel-wise versus single channel in 2-D space with t-SNE.

Second, we visually assessed the performance difference in MAPE between the channel-wise and single-channel methods based on patient journey characteristics. The diversity of clinical codes and the portion of claim events capture the heterogeneous levels of the patients' journeys. As can be seen, the majority of the relatively healthy patients with the least need severity (intense blue color) have a large portion of their claim events (circle size) with low clinical code diversity, so the performance difference is not significant for this group (light red color). On the other hand, the majority of the disabled patients with the highest need severity (intense blue color) have a large portion of their claim events (circle size) with high diversity, and thus the MAPE difference is the greatest for this group of patients (intense red color). In addition, a similar experiment was conducted to explore additional attributes of the patient journey by utilizing the distinct number of clinical codes instead of their diversity, and the number of providers rather than claim events. This analysis is detailed in online Appendix B and yielded comparable observations. As a result, the proposed method becomes more effective as the heterogeneity of the patients' journeys increases.

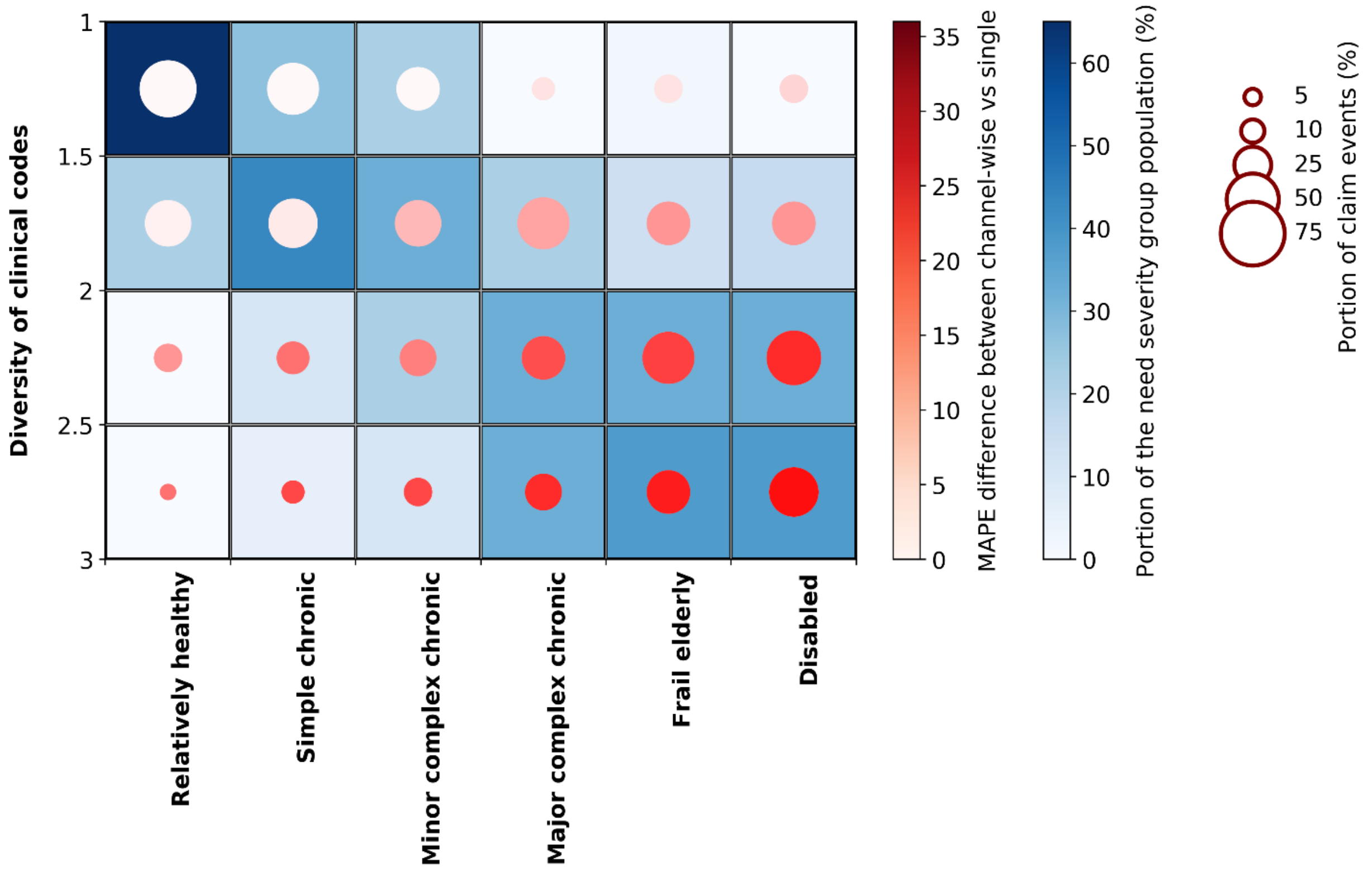

Figure 4. MAPE difference between channel-wise versus single channel based on patients' journey characteristics, including the diversity of clinical codes and the portion of claim events.

Third, to better understand the underlying attention mechanism that allows the channel-wise architecture to pick up on codes across three channels for HN patients, we depicted the feature importance of a rheumatoid arthritis (RA) profile using both architectures. To archive this, we used the criteria proposed in (Chung et al. 2013, Widdifield et al. 2013) to detect RA patients and their lines of therapy. Additionally, we added an attention layer to the single-channel architecture to allow for apples-to-apples comparisons. Figure 5 shows an example of the last 20 claim events of a RA patient with an HN patient profile (i.e., major complex chronic), in which costs increased by approximately 85% in the following year (result period) compared to the previous year (observation period). One of the primary reasons was that the patient moved from first-line therapy (1st LT) to second-line therapy (2nd LT) the following year. In 2016, the average annual direct medication cost for a RA patient in the 1st LT was $717, compared to $35,896 for those in the 2nd LT in the United States (Khanna and Smith 2007, Muszbek et al. 2019). A similar significant cost difference exists between the two lines of therapies worldwide (Chen et al. 2019). T-1 is the claim event day that this patient was moved from the 1st LT to the 2nd LT, and eight providers were seen on that day. As seen in Figure 5, due to this large number of same-day visits, the single-channel method missed identifying the importance of the 2nd LT medication and diagnosis, while the channel-wise method accurately identified the importance. As a result, this patient's Absolute Percentage Error (APE) with the channel-wise method is 70% lower than that of the single channel. This is an example use case, in which the proposed method detects codes across all three channels in ways that are more appropriate for higher-need groups. Figure B2 in online Appendix B contains another example.

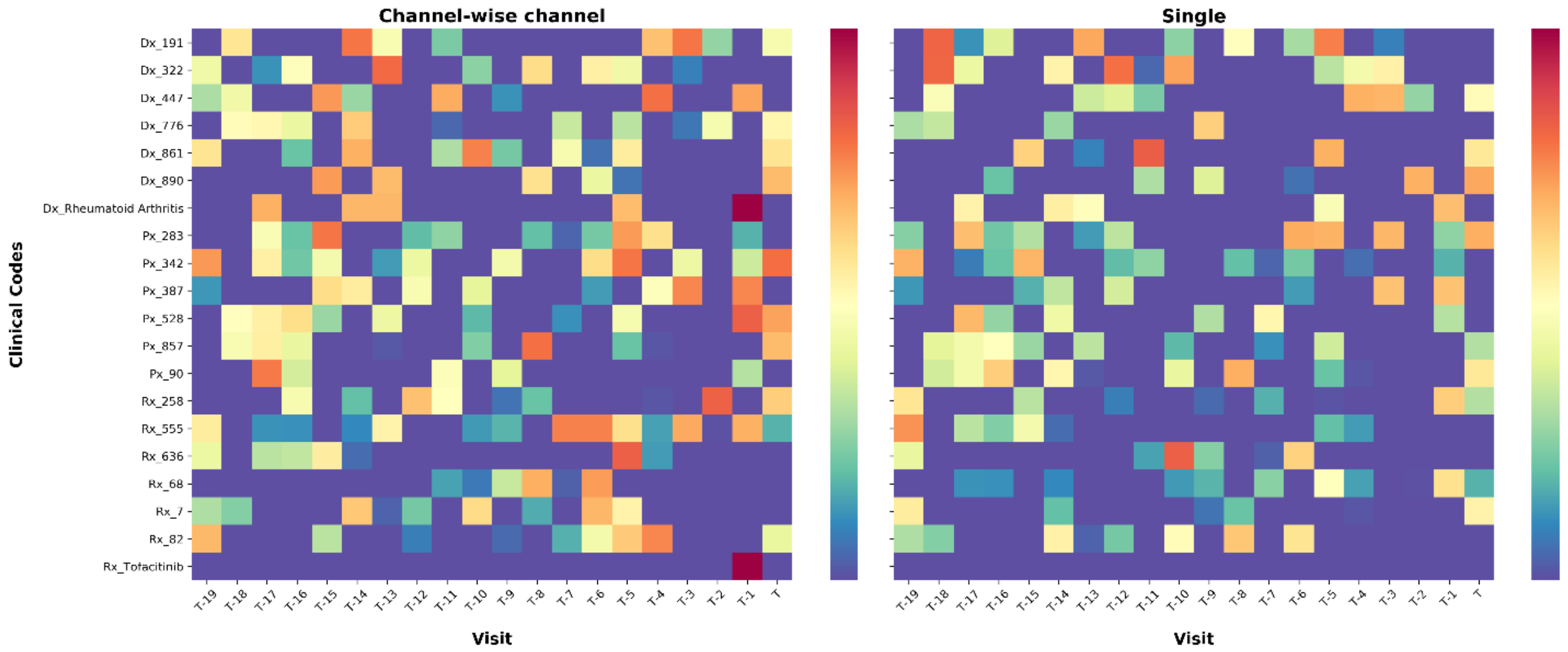

Figure 5. Example of features importance, detected from attention weights, for the last 20 claim events (T) of a RA patient. In order to identify RA patients and their line of therapy, we used the following criteria proposed in (Chung et al. 2013, Widdifield et al. 2013), where *Tofacitinib* is identified as one of the medications used in the 2nd LT. It should be noted that—because it is impossible to display the thousands of diagnosis, medication, and procedure codes on each patient profile—this figure only depicts a small portion of the patient's profile, where the two RA-related diagnosis and medication codes are pronounced.

Finally, to visually shed light on the prediction process of the proposed method, we explored some examples of attention weights for different types of patients with various levels of need severity, and this information can be found in online Appendix B.

## 5.4. Ablation analysis

We used an ablation analysis to illustrate the effect of each component of the proposed channel-wise learning method. Specifically, comparisons included (I) single-channel representation learning with unilateral bag-of-codes for all types of medical codes versus multi-channel representation learning with separate bag-of-codes for each type of medical code, (II) pre-trained embeddings with Doc2Vec versus trainable embeddings, and (III) the combination of the separate channels with attention versus the simple concatenation of channel output, which are all separately evaluated in this experiment.

Table 7 shows the results of an ablation analysis with total claims cost as the target variable and MAPE as the performance criterion. One initial observation is that, regardless of the embedding method, employing channel-wise representation learning with separate bag-of-code types is more effective than a single-channel approach with a bag-of-all-code-types ($p<0.01$). This advantage is more pronounced for HN patients compared to healthier patients. The single-channel approach, originally proposed for single source data collected by healthcare providers, works best in settings where various types of medical codes are unambiguously attached to a single visit, thereby providing a guarantee to their relationship. Mixing all code categories together, without taking their unique natures into account, is therefore not effective for AC data, given its finest granularity is not at the visit level. Moreover, pre-trained embeddings on medical events with Doc2Vec are more effective than trainable embeddings for AC data, regardless of whether single or multiple channels are utilized. Although this advantage is not significant for relatively healthy patients, it is almost consistent over all need severity categories. This aligns with previous studies which indeed showed that pre-trained embeddings on medical codes with Word2Vec were more effective than trainable embeddings (Choi et al. 2017, Choi, Bahadori, Schuetz, et al. 2016) (Section 2.2). Finally, we note that combining individual RNN channels with an attention layer is more effective than a simple concatenation of their outputs, since the attention layer can capture interconnectivity among the extracted patterns from each channel. This is particular to channel-wise representation learning, since the attention layer contribution with single-channel representation learning is minor.

Table 7. Ablation study for identifying the specific contributions of various components to MAPE prediction performance over different need severity categories.

| Learning | EmdApp | Att | All | Relatively healthy | Simple chronic | Minor complex chronic | Major complex chronic | Frail elderly | Disabled |
|---|---|---|---|---|---|---|---|---|---|
| Single channel | Trainable | N | 62.9 | 31.4 | 37.7 | 50.2 | 70.9 | 93.5 | 117.2 |
| | Trainable | Y | 62.6 | 31.3 | 37.4 | 49.9 | 70.4 | 93.2 | 116.9 |
| | Doc2Vec | N | 61.3 | 31.1 | 36.1 | 48.4 | 68.5 | 92.3 | 115.4 |
| | Doc2Vec | Y | 60.8 | 30.9 | 35.7 | 47.9 | 68.1 | 91.8 | 114.7 |
| Channel-wise | Trainable | N | 54.2 | 30.4 | 35.1 | 44.6 | 57.9 | 77.6 | 99.4 |
| | Trainable | Y | 52.1 | 29.8 | 34.3 | 42.6 | 55.1 | 74.2 | 96.1 |
| | Doc2Vec | N | 49.2 | 29.5 | 33.2 | 41.5 | 49.4 | 69.2 | 90.7 |
| | **Doc2Vec** | **Y** | **46.8** | **28.9** | **30.9** | **39.1** | **46.9** | **66.2** | **87.5** |

**EmdApp**: Embedding approach. **Att**: Attention model.

## 5.5. Monetary analysis

After selecting the best architecture for AC-based cost prediction from a technical perspective, we investigated the added monetary value of the proposed method compared to the most competitive benchmarks from previous experiments. In particular, we implemented single-channel versus multi-channel architectures, pre-trained versus trainable embedding, and successful benchmarks from previous experiments as additional benchmark measures. Also, pursuant to the benefit of the proposed method for heterogeneous profiles of HN patients, we further conducted the monetary analysis for all HN groups.

Figure 6 depicts the MAE performance of various network architectures. Compared to single-channel architectures, the proposed channel-wise learning technique can reduce cost prediction error by up to $5,000 ($31K vs. $36K) for disabled patients, and its advantage over other literature benchmarks is similar.

Figure 7 depicts the net value in millions of dollars (as well as total-over and total-under values in millions of dollars) of cost prediction errors for the same set of architectures shown in Figure 6. Comparing single channel with trainable embedding versus the proposed channel-wise RNN method, $239 million ($1,091 vs. $1,330) worth of prediction error can be mitigated by the avoidance of overpaying $106 million ($538 vs. $644) and underpaying $133 million ($553 vs. $686). This underscores the managerial

implications of the current work, in that it helps insurers to more accurately allocate resources, provides health plan providers with proper compensations, and directs patients to a more well-informed decision-making process regarding their future healthcare.

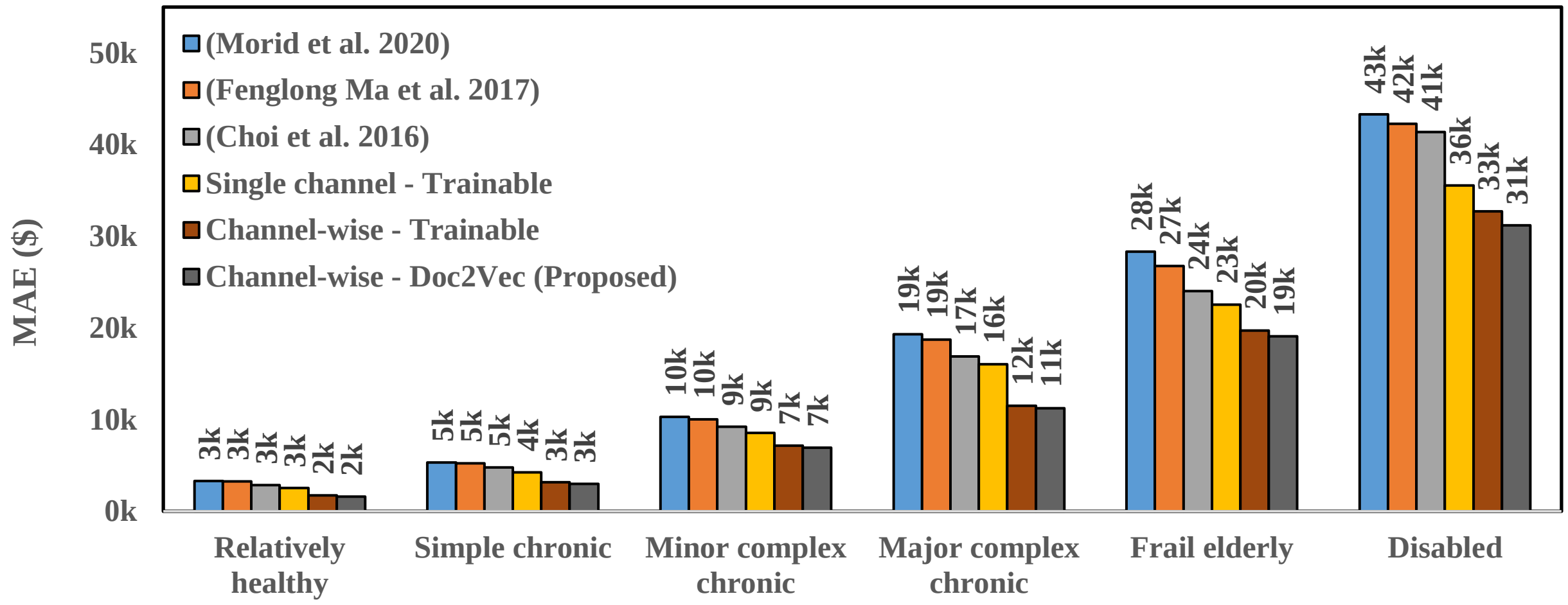

Figure 6. MAE for the dollar value of total claims cost prediction under different deep learning architectures over need severity categories.

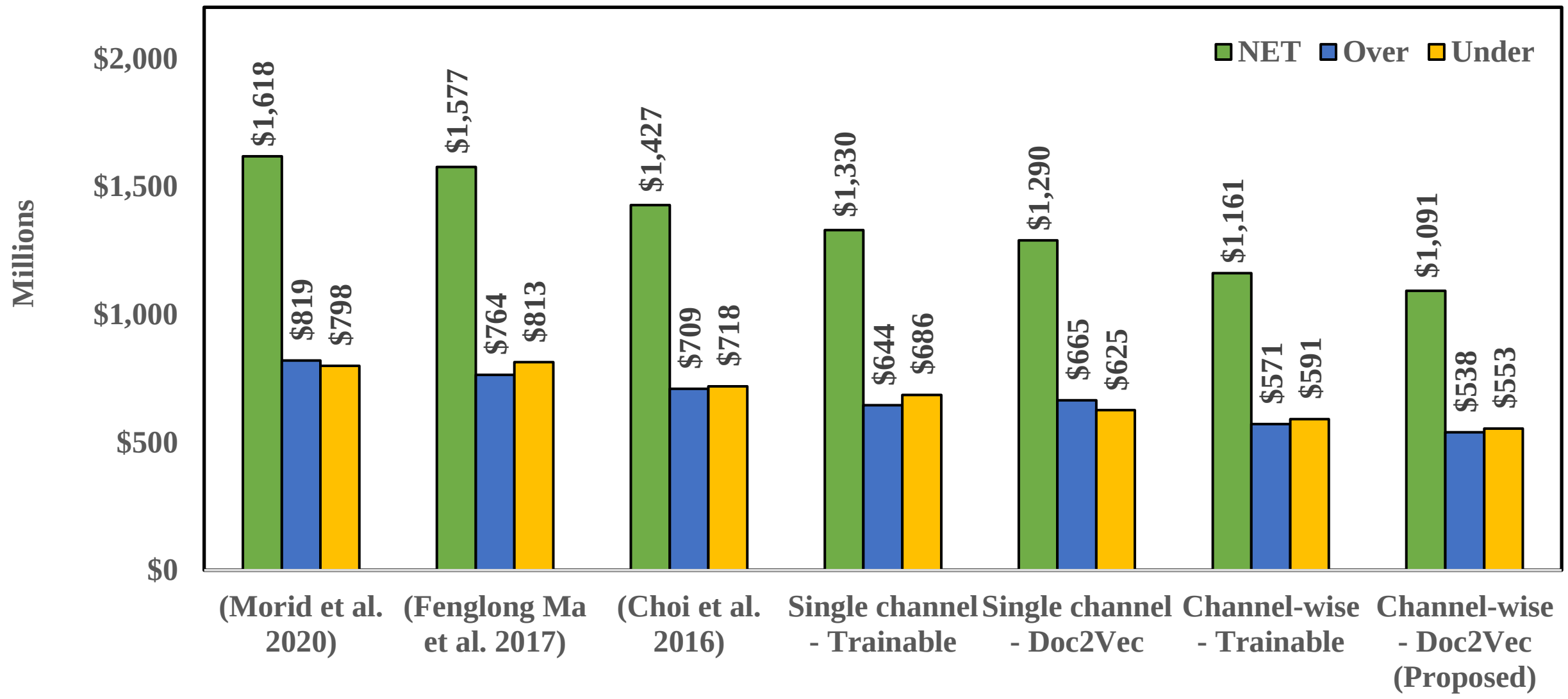

Figure 7. Net pay, underpayment, and overpayment values, in millions of dollars, for total claims cost prediction errors mitigated under different deep learning architectures.

## 5.6. Sensitivity analysis

To demonstrate the sensitivity of the proposed model to selected architectural settings, we investigated the performance using monthly or weekly aggregation as opposed to raw daily medical events, and we also assigned one channel per medical code as opposed to code type. Additionally, we compared the number of parameters for each architecture variation. It should be mentioned here that we investigated embedding with transformer architectures, but the large number of parameters and dependence on medical-code ordering prevented them from achieving relatively successful performances. The results of this experiment can be found in Appendix C.

### 5.7. External validation

For an external validation of our methods, we used a secondary dataset consisting of approximately 8.5 million medical claims and 2.4 million pharmacy claims from 134,000 Medicare patients residing in California from January 2017 to December 2020. The same ablation analysis experiments that were performed on the primary data were also employed on this external dataset. Moreover, to evaluate the transferability of the knowledge extracted from one AC system to another (see Section 2.2), we imported and compared the pre-trained embedding vectors from the Utah dataset (i.e., the internal validation) against learning all the embedding vectors for the California dataset with a trainable network layer. The results of this experiment can be found in Appendix D.

## 6. Multi-channel Entropy Index for Flexible Heterogeneous Patient Evaluation

To leverage the link between patient heterogeneity and data heterogeneity for the flexible evaluation of cost prediction models for heterogeneous patients, we define a multi-channel entropy index to capture the heterogeneity in a claim event, *E*, making accommodations for both the number of associated codes as well as their diversity:

$$Entropy\ (E) = -\ Len(E)\ \times [P(E_{dx}) \times Log\big(P(E_{dx})\big) + P\big(E_{px}\big) \times Log\ \Big(P\big(E_{px}\big)\Big) +\ P(E_{rx}) \times Log\big(P(E_{rx})\big)]$$

(11)

where $E_{dx}$, $E_{px}$, and $E_{rx}$ are the total number of diagnosis codes, procedure codes, and medication codes, respectively. Also, *Len (E)* is the total number of clinical codes, while *P(X)* and *Log(P(X))* respectively represent the probability value and logarithmic probability value of each type of clinical code. In order to avoid *Log(0)*, all codes are increased by one. Examples of claim events with different *Entropy* can be found in Appendix H. The entropy of a patient profile is then calculated as a simple average over the number of claim events, $T$, listed for that patient:

$$ProfileEntropy = \frac{\sum_{i=1}^{T} Entropy\,(E_i)}{T} \qquad (12)$$

The potential of this profile entropy for the flexible evaluation of heterogeneous patients hinges on its close alignment to the levels and characteristics of heterogeneous patient needs. We empirically probe into this relationship in the following ways. First, we compare the average normalized profile entropy (i.e., *ProfileEntropy* in (12) is scaled between 0 and 1) of patients belonging to each of the six severity categories in Table 8. Because the volume and diversity of the codes in patients' claim events increase over patient severity levels and the coarseness of data granularity, the average profile entropy also increases over these two dimensions.

Table 8. Average *ProfileEntropy* of patients in different need levels

| | **Granularity** | **Relatively healthy** | **Simple chronic** | **Minor complex chronic** | **Major complex chronic** | **Frail elderly** | **Disabled** |
|---|---|---|---|---|---|---|---|
| Average entropy | Day | 0.09 | 0.17 | 0.29 | 0.35 | 0.41 | 0.52 |
| | Week | 0.10 | 0.19 | 0.32 | 0.42 | 0.49 | 0.58 |
| | Month | 0.11 | 0.22 | 0.39 | 0.55 | 0.64 | 0.82 |

Second, we derive and present the Pearson correlation coefficients among patients' profile entropy values, and the values of the characteristics of patient heterogeneity in Table E1 in Appendix E. As expected, high correlations—around 0.7—between patients' profile entropy and their volume of codes, claim events, or providers are instrumental in establishing a strong alignment between patients' profile entropy and severity levels. Last, we examine if this profile entropy can also robustly confirm the link between data heterogeneity and patient heterogeneity based on another patient categorization—high-cost patients (i.e., the top 5% of patients according to average annual cost) versus low-cost patients (the remaining 95% of patients). Table E2's third and fourth columns in Appendix E show that the complexity of patient journeys for high-cost patients (e.g., the volume of providers, claim events, and codes average at 5.86, 32.23, and 36.77 per month, respectively) is significantly higher than that of the low-cost patients (e.g., averaging at 4.29 providers, 22.63 claim events, and 17.54 codes per month per patient, respectively).

The distinct levels of average profile entropy between these two populations (e.g., 0.41 versus 0.84 at the month level) indicate that these data heterogeneity metrics are closely aligned with patient heterogeneity based on cost categorization as well. The results from this probe suggest that *ProfileEntropy* helps identify the link between patient heterogeneity and data heterogeneity, providing the potential for the use of a data heterogeneity metric that is closely aligned with patient heterogeneity in order to broaden the categorization and evaluation of heterogeneous patients. In this study, we examine this potential for two scenarios using *ProfileEntropy* to delineate heterogeneous patients.

In one of the scenarios, we assume that the distributions of patient heterogeneity subpopulations (e.g., need-based) are known from prior studies. By segmenting patients based on a given distribution using *ProfileEntropy*, a decision-making party could avoid the codification bottleneck or gain a quick assessment of the prediction options for the targeted groups of patients before embarking on the resource-demanding process of codifying patients by need levels or another criterion. Table E2 compares the characteristics of patient heterogeneity and improvements in MAPE between codification-based categorization and entropy-based categorization according to the distributions for the two need levels (i.e., high need and low need) and two cost levels (i.e., high cost and low cost) derived from the study's main dataset. The results reveal that the significant performance improvements made by the proposed channel-wise cost prediction framework for HN or high-cost patients are consistently present in the corresponding high-entropy groups. Such applications of the proposed *ProfileEntropy* help stakeholders of cost prediction implementation and analysis to efficiently and effectively accomplish the task of selecting a model that benefits HN or high-cost patients more than others.

In the other scenario, we assume that the distributions of subpopulations of heterogeneous patients are unknown or relatively dynamic. In this case, stakeholders could benefit from stratifying heterogeneous patients into strata (e.g., of equal size in the simplest case) using *ProfileEntropy* for flexible exploration and analysis. For illustration purposes, Table 9 displays patient characteristics and cost prediction model comparisons among the quintiles of *ProfileEntropy*. The comparisons confirm the expected data heterogeneity and patient heterogeneity alignments as well as the high performance improvements for patients in the high ranges of *ProfileEntropy*. Since the stakeholders in this scenario might want to drill into performance comparisons with ablation and financial analyses, we present such comparisons in Table E3 and Figure E1 in Appendix E. These analyses demonstrate the practical utility of utilizing a data heterogeneity measurement that is well aligned with patient heterogeneity characteristics in order to help flexibly examine the degree to which a predictive model can help reduce healthcare disparity.

Table 9. Distribution of chronic disease and total MAPE performance over entropy buckets.

| | | **Quintiles *of ProfileEntropy*** | | | | |
|---|---|---|---|---|---|---|
| | | **0-20** | **20-40** | **40-60** | **60-80** | **80-100** |
| Chronic | | 12% | 18% | 22% | 29% | 38% |
| Average number of providers | Day | 1.89 | 2.72 | 3.88 | 4.83 | 7.81 |
| | Week | 1.96 | 2.88 | 4.02 | 5.44 | 8.74 |
| | Month | 2.19 | 3.23 | 4.68 | 7.03 | 10.96 |
| **Learning** | | | | | | |
| Single channel | | 39.1 | 44.9 | 58.6 | 76.8 | 94.2 |
| Channel-wise | | 32.3 | 35.6 | 43.8 | 53.7 | 66.6 |
| **Improvement (Difference)** | | **6.80** | **9.30** | **14.80** | **23.10** | **27.60** |

# 7. Conclusion

This study sheds light on the socio-technical considerations underlying the alignment of AC data heterogeneity with patient heterogeneity, offering a new channel-wise cost prediction framework to reduce AC data heterogeneity and prediction errors, as well as a multi-channel entropy index to foster the investigation of patient heterogeneity in impact evaluations that are essential in helping to reduce healthcare payment disparities. Our research adheres to the “three ‘I’ principles” of interestingness, impact, and integration expected of the design-insight abstraction (Abbasi et al. 2024). The review of related work and research gap discussion demonstrate ways in which this work not only integrates but also extends previous research. Beyond the cost prediction application, the proposed architecture and evaluation designs hold significant potential for problem domains exhibiting similar mappings of social considerations into heterogeneity in multi-source data (e.g., online consumer data on product research, sales, promotions, or fulfillment and return in omni-channels of different commerce, advertisement, and social media platforms).

Related to the interesting principle, the evaluation results provide interesting and strong empirical support for the attainment of this work's guiding social and technical desiderata. In particular, the channel-wise architecture, the attention mechanism, and the pre-trained PV-DBOW embedding demonstrate super-additive synergies to help reduce the adverse effect of data heterogeneity for cost prediction. The proposed architecture also enhances patient representation coherence (see Figure 3) in the same severity category and the importance differentiability of temporal and medical code signals (see Figures 5 in Section 5.3 and B1 in Appendix B). While the embedding and attention mechanism can lift the performance of the single-channel benchmark, the channel-wise architecture achieves greater utility from these architectural components. The proposed architecture pushes the boundary of extant cost-prediction and other non-cost health outcome deep learning models to offer much-needed accurate predictions for HN patients such that they can experience more significant reductions in insurance payment discrepancies. Our analysis confirms high correlations between the proposed multi-channel entropy index and patients' characteristics, enabling an alternative to approximate heterogeneous patient need strata using this data heterogeneity measurement, which thus allows for flexible and scalable evaluations of decision support.

The evidence of the impact principle stems from the positive impact of the design and evaluation insights for reducing healthcare disparities among HN patients via both the domain-knowledge-based and the alternative data-heterogeneity-based delineation of patient subpopulations as well as this work's implications for both healthcare and non-healthcare audiences.

For healthcare audiences, the proposed method can benefit private healthcare payers such as government agencies (e.g., CMS) by improving the distribution of administrative costs through better resource allocation. More specifically, current RAMs in CP programs, such as MA plans, can benefit from this study because they rely heavily on accurate healthcare cost forecasting to ensure that there are adequate resources to reimburse health plans for patient services. This also benefits health plans by mitigating underpayments for HN patients and diminishing the incentives that result in health plans favoring healthy patients over HN patients. Furthermore, the multi-channel entropy index provides a computationally efficient alternative for healthcare practitioners and researchers working with HN patients. It allows for the simple extraction of patient profiles with different levels of heterogeneity, as opposed to relying on expert-driven definitions that require extensive domain knowledge and the time-consuming extraction of profiles based on hundreds of diagnosis and procedure codes. Finally, for healthcare researchers and practitioners working with AC data, the proposed channel-wise representation learning pipeline can aid in reducing heterogeneous data complexity and facilitating data processing for the downstream machine learning application.

For non-healthcare audiences, this study suggests integrating humanistic outcomes, such as disparities in economic or operational elements, in the evaluation design process. Furthermore, the multi-channel entropy index, which identifies different levels of heterogeneity profiles, can be used in other applications where subjects also embark on multi-source temporal heterogeneous journeys. For such applications, channel-wise ML design can help reduce the heterogeneity of multi-source temporal data as well.

The extent to which the proposed architecture and data heterogeneity metric can benefit the underlying applications varies along two dimensions—the business outcome disparity and the alignment between data heterogeneity and the heterogeneity exhibited in individuals' related journeys. Figure F1 in Appendix F depicts four quadrants of varying benefit potentials according to the levels of disparity and heterogeneity alignment. Because this study has been motivated by the need to reduce high business outcome disparity by leveraging high patient-data heterogeneity alignment in quadrant 1 (i.e., Q1), we believe that the benefits of applying the proposed prediction framework and *ProfileEntropy* would also be the highest in similar application contexts. In quadrant 2 (Q2), when outcome disparity is high but the alignment between data heterogeneity and an individual's heterogeneous journeys is low, the benefits of this study's architectural and evaluative insights could still be significant enough to justify their adoption. The benefits and therefore the motivation for applying the proposed prediction architecture and evaluation metric cannot sustain in quadrants 3 and 4 (Q3 and Q4), as they lack a sufficiently high alignment between the characteristics of heterogeneous individuals and data heterogeneity.

There are several ways in which this study could be furthered in future work. First, while accurate RAMs can significantly help payers to assure that health plans receive proper compensation per patient (Kautter et al. 2012) and reduce incentives fueling the biased enrollment of low-risk patients and the disenrollment of HN patients (Schone and Brown 2013), the problem is not fully resolved. An accurate model provides a strong tool for the payers as well as the insurance plans at the same time, which can still lead to the biased selection of patients by insurance plans. While the initial purpose of this study was to find an accurate artificial intelligence (AI) model as the first step in solving this problem, future studies are needed to investigate optimal governance policies that will lead to the responsible use of AI. Second, a primary goal of this paper was to demonstrate the benefits of channel-wise learning. As such, we did not leverage the effects of static variables such as age, gender, and ethnicity on prediction performance. Since numerous approaches have been suggested for combining static and dynamic features for deep time series prediction tasks in the healthcare domain (Morid et al. 2022), finding an optimal architecture for channel-wise learning that includes both static and dynamic input features holds great promise for future exploration. Third, we did not have access to patient EHR records in different hospitals, and thus we could not explore additional clinical features that could potentially enhance cost prediction performance. Although linking EHR and AC data is an arduous task, we encourage its exploration in future research studies. Fourth, future studies measuring multi-source data heterogeneity for problem domains with different data heterogeneity challenges may consider exploring additional functional forms of diversity measurements.

## Acknowledgment

We would like to express our heartfelt appreciation to the editors and anonymous reviewers for their insightful feedback and constructive suggestions throughout the review process. Their guidance played a crucial role in enhancing the quality, rigor, and clarity of this paper.

# Appendix A: Severity of Patients' Needs

Table A1 shows the list of diseases and indicators used to define the various severity of patients' needs according to (Joynt et al. 2017). A detailed list of the procedure codes and diagnosis codes used to identify each condition or indicator can be found in the same reference.

Table A1 shows the list of the non-complex and complex chronic conditions, as well as frailty indicators.

| **Complex chronic conditions** | **Non-complex chronic conditions** | | **Frailty indicators** |
|---|---|---|---|
| Acute MI / Ischemic heart disease | Amputation status | Immune disorders | Abnormality of gait |
| Chronic kidney disease | Arthritis and other inflammatory tissue disease | Hyperlipidemia | Protein-calorie malnutrition |
| Congestive heart failure | Artificial openings | Liver and biliary disease | Adult failure to thrive |
| Diabetes | Benign prostatic hyperplasia | Cancer | Cachexia |
| Dementia | Neuromuscular disease | Osteoporosis | Debility |
| Lung disease | Cystic fibrosis | Paralytic diseases | Difficulty in walking |
| Psychiatric disease | Endocrine and metabolic disorders | Skin ulcer | Fall |
| Specified heart arrhythmias | Eye disease | Substance abuse | Muscular wasting and disuse atrophy |
| Stroke | Hematological disease | Thyroid disease | Muscle weakness |
| | Inflammatory bowel disease | Hypertension | Decubitus ulcer of skin |
| | | | Senility without mention of psychosis |
| | | | Durable medical equipment (cane, walker, bath equipment, and commode) |

# Appendix B: High-need Analysis

Figure B1 depicts the MAPE performance difference between the channel-wise and single-channel methods based on various patient journey characteristics, including the unique number of clinical codes and the number of providers. As can be seen, the majority of the relatively healthy patients with the least need severity (intense green color) visit a small number of providers (circle size) with a small number of clinical codes, so the performance difference is not significant for this group (light pink color). On the other hand, the majority of disabled patients with the highest need severity (intense green color) visit a large number of providers (circle size) with a large number of clinical codes, and thus the MAPE difference is the greatest for this group of patients (intense pink color).

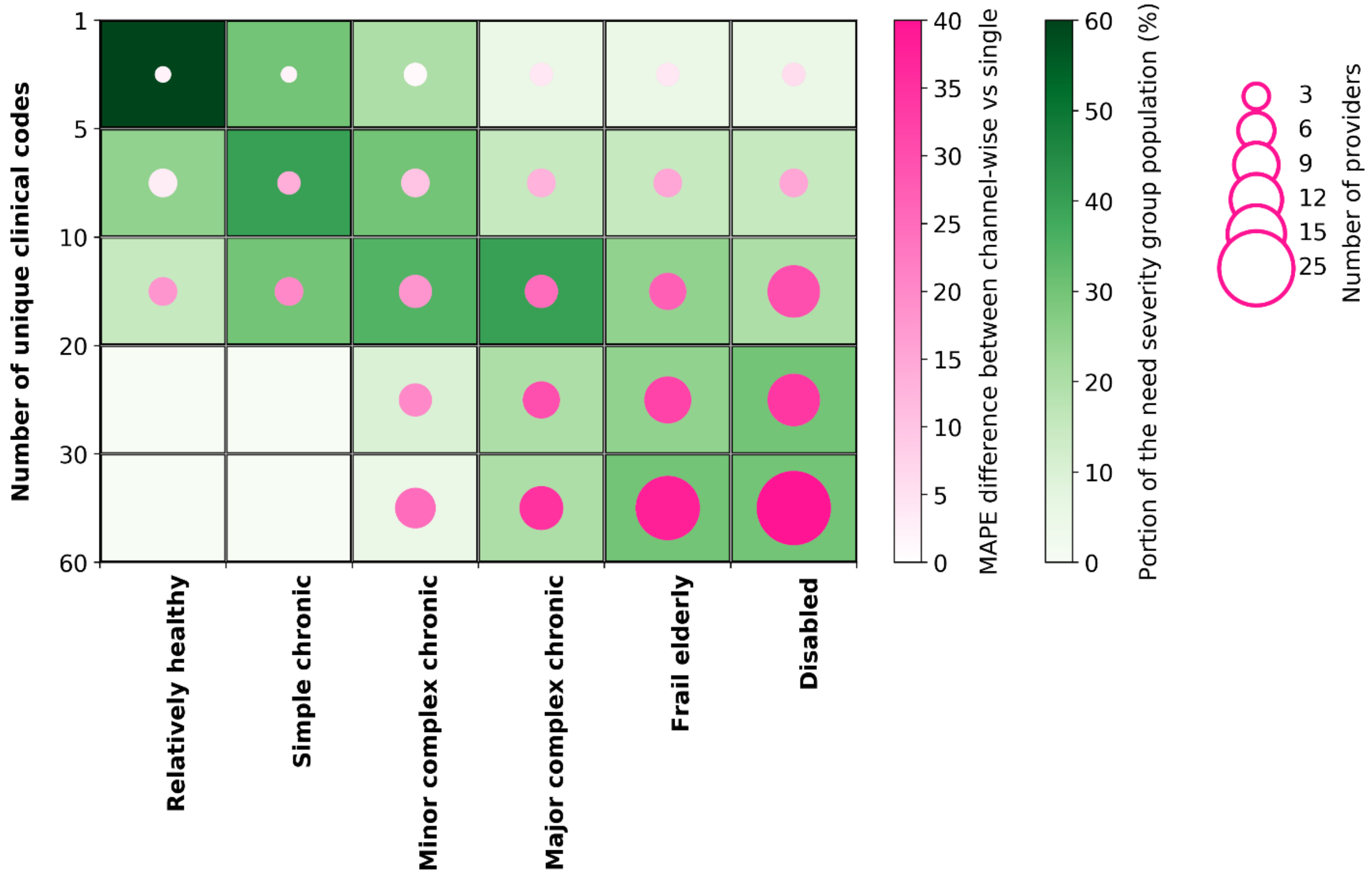

Figure B1. MAPE difference between channel-wise versus single channel based on patients' journey characteristics, including the diversity of clinical codes and the number of claim events.

Figure B2 shows an example of the last 20 claim events of a diabetic patient with a HN patient profile in the frail elderly group, in which costs increased by approximately 32% in the following year (result period) compared to the previous year (observation period). One of the primary reasons was that the patient's diabetes led to stroke complications, which contributed to more than 72% of the increased cost. It has also been reported that strokes are one of the top three most expensive complications of diabetes (Yang et al. 2020). On T-3 and T-2 claim event days, this patient was diagnosed with a stroke as a complication of diabetes. While both the channel-wise and single-channel methods have identified the diagnosis, the former method gave considerably more attention to it. As a result, this patient's APE with the channel-wise method is 33% lower than that of the single channel.

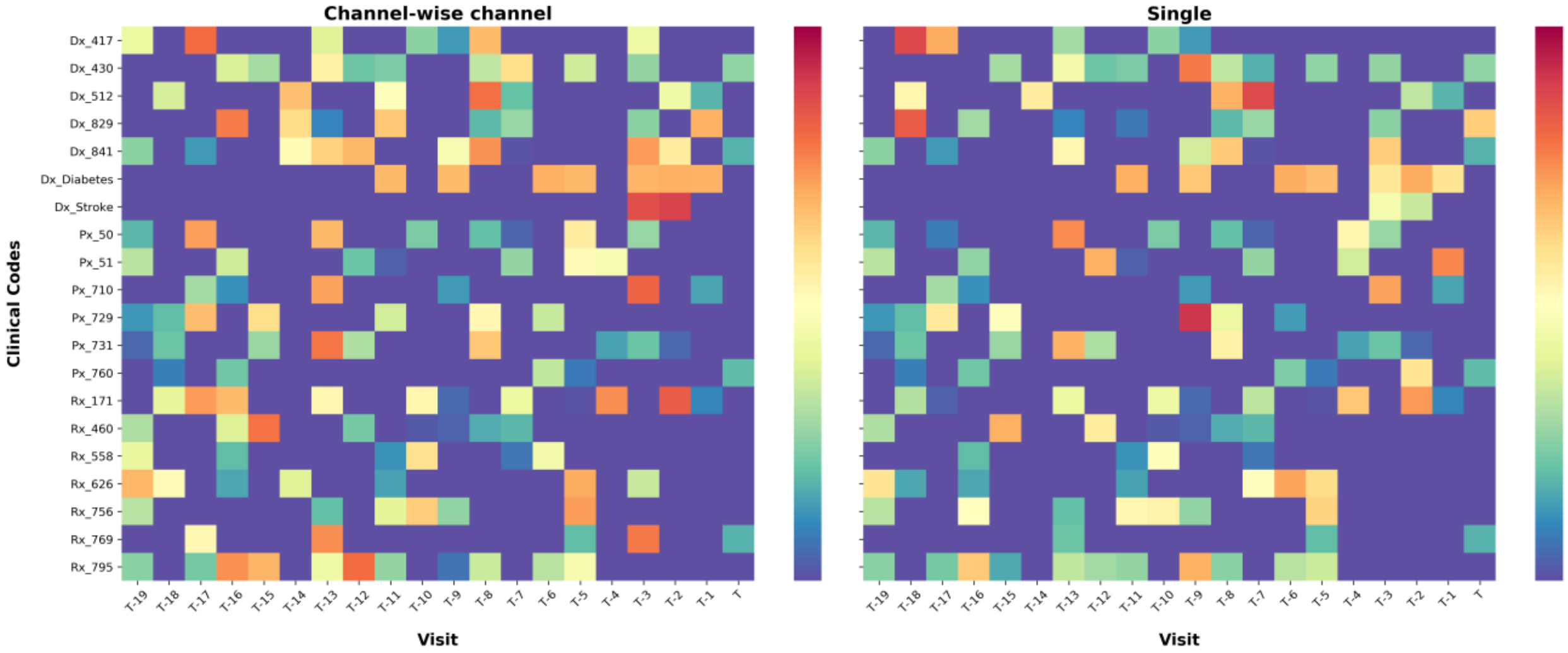

Figure B2. Example of features importance, detected from attention weights, for the last 20 claim events (T) of a HN patient. It should be noted that—because it is impossible to display the thousands of diagnoses, medications, and procedure codes on each patient profile—this figure only depicts a small portion of the patient's profile, where the two related diagnosis are pronounced.

Figure B3 below shows the relative values of attention weights for the last ten medical events (horizontal axis) occurring with four different patients (vertical axis). The colored spectrum of attention scores learned by the attention mechanism varies according to the type of diagnosis, as well as the costs associated with the diagnosis. Specifically, Patient #1 is in the "relatively healthy" need severity category; Patient #2 has been diagnosed with a chronic disease between T-2 to T-5 and is moving to the "Simple Chronic" need severity category; Patient #3 has been diagnosed with a complex chronic disease between T-5 to T and is moving to the "Major Complex Chronic" need severity category; and Patient #4 is in the "frail elderly" need severity category, having at least two medical providers and one pharmaceutical provider on all ten medical events (except T-1 and T-8). These simple visualizations of the prediction model's differential reliance on the claims data over different days in a year for different patient categories help elucidate how the proposed method assigns claims costs, which can facilitate its adoption for practical integration in healthcare systems. For example, the attention weights over time are consistently higher for an elderly frail patient than for a relatively healthy individual. Chronic disease diagnoses prompt the model to place higher attention weights on the claims submitted for the subsequent visits, which last longer for more complex chronic diseases.

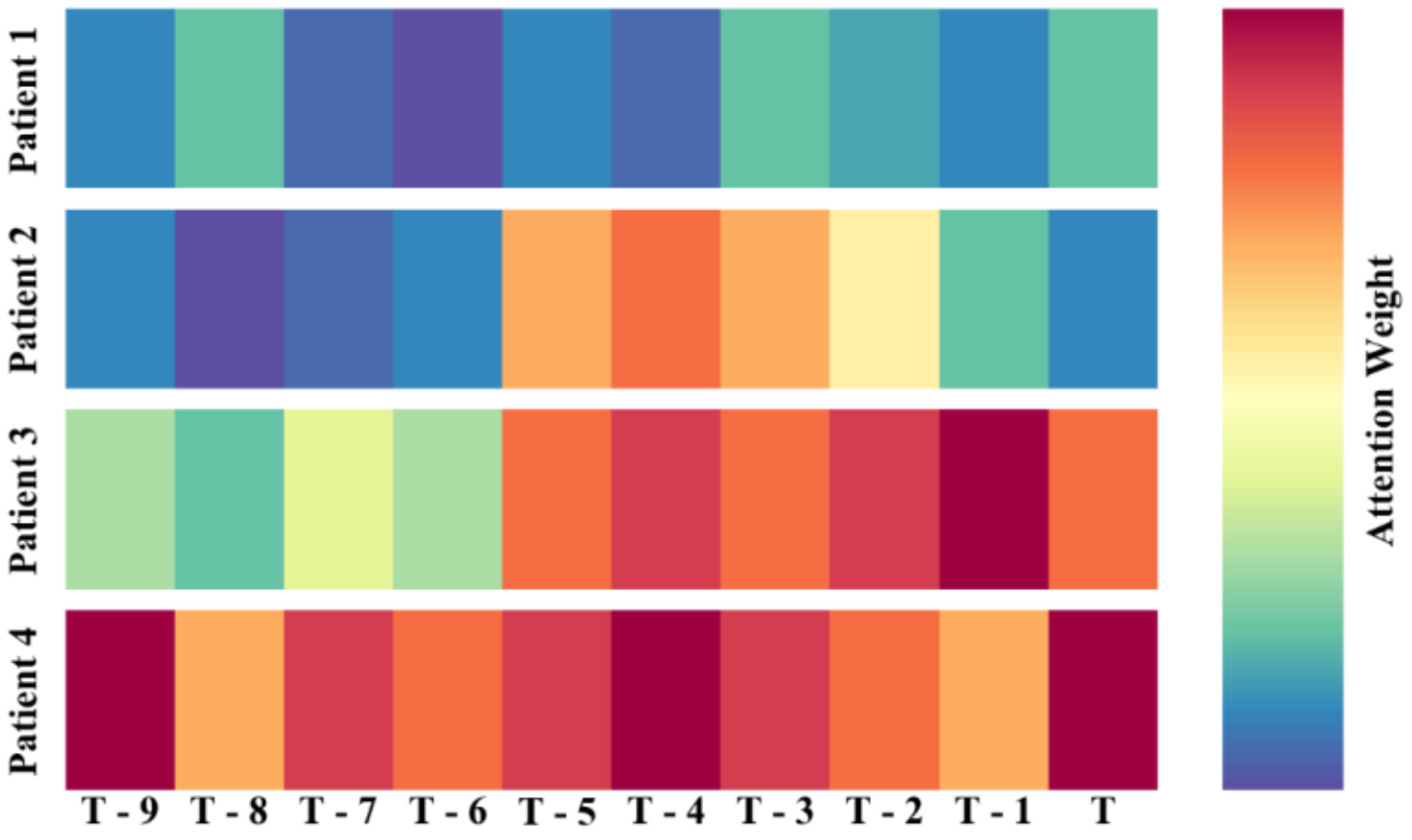

Figure B3. Examples of attention weights generated for patients in different need severity categories.

# Appendix C: Sensitivity Analysis

Table C1 shows the incremental performance of the proposed channel-wise architecture versus the single channel by adjusting various settings. Regardless of the leaning approach, coarser temporal granularities reduce performance by aggregating more data from heterogeneous patient profiles. Second, training one channel per medical code does not assist accuracy, due to the sparsity inherent to medical codes. Third, regardless of the learning approach, pre-trained embeddings both improve accuracy and reduce the number of required network parameters. Since the pre-trained embeddings are stored in a look-up matrix, there is no need for training an additional layer, dramatically decreasing the network size. As mentioned in Section 5.4, this finding echoes previous EHR studies within extant literature demonstrating that pre-trained medical code embeddings with Word2Vec are more effective than trainable embedding (Choi et al. 2017, Choi, Bahadori, Schuetz, et al. 2016).

Table C1. Sensitivity analysis of total claims cost prediction.

| Learning approach | Temporal granularity (MAPE) | | |
|---|---|---|---|
| | **Month** | **Week** | **Day** |
| Single channel | 68.9 | 65.0 | 62.9 |
| Channel-wise | 55.2 | 50.6 | 46.8 |
| | **Channel choice (MAPE)** | | |

| | One channel per code | One channel per code type | |
|---|---|---|---|
| Channel-wise | 75.7 | 46.8 | |
| | **Representation learning (MAPE)** | | |
| | Trainable | Pre-trained (Doc2Vec) | |
| Single channel | 62.6 | 60.8 | |
| Channel-wise | 52.1 | 46.8 | |
| | **Representation learning (Number of parameters)** | | |
| | Trainable | Pre-trained (Doc2Vec) | |
| Single channel | 2,581,377 | 582,145 | |
| Channel-wise | 6,927,238 | 929,542 | |

# Appendix D: External Validation

Table D1 summarizes model performance on an external validation dataset with an additional ablation analysis. Similar patterns to those found with internal validation experiments were observed, again confirming the robustness of the proposed architecture.

We did not observe a significant difference between using the pre-trained embedding vectors from the Utah dataset for the external California dataset and learning all the embedding vectors for the California dataset with a trainable network layer ($p > 0.351$). As such, only one set of results are reported in Table D1. This effect is likely due to the size of the external validation dataset, which is large enough for a BiGRU model to reliably train custom embedding vectors. However, trainable embedding is time-consuming, and the feasibility of obtaining reliable representations depends on the availability of infrastructural resources (Zhuang et al. 2021). Importantly, this research study provides evidence that the knowledge learned from the embedding vectors of one AC dataset is transferable to another.

Table D1. External validation of the ablation study for cost prediction.

| **Learning** | **EmdApp** | **Att** | **All** | **Relatively healthy** | **Simple chronic** | **Minor complex chronic** | **Major complex chronic** | **Frail elderly** | **Disabled** |
|---|---|---|---|---|---|---|---|---|---|
| Single channel | Trainable | Y | 53.3 | 20.9 | 32.7 | 41.3 | 60.2 | 79.4 | 102.7 |
| | Doc2Vec | Y | 51.6 | 20.3 | 31.5 | 39.4 | 58.1 | 77.6 | 100.5 |
| Channel-wise | Trainable | N | 45.0 | 19.9 | 30.1 | 33.3 | 49.2 | 67.2 | 88.4 |
| | Trainable | Y | 41.8 | 18.8 | 27.1 | 30.2 | 45.4 | 63.4 | 84.9 |
| | Doc2Vec | N | 38.5 | 16.4 | 24.3 | 27.1 | 42.6 | 59.8 | 78.3 |
| | **Doc2Vec** | **Y** | **34.7** | **14.8** | **22.1** | **24.2** | **37.2** | **54.4** | **72.1** |

# Appendix E: Analyses Based on Multi-channel Entropy Index

Table E1. Pearson correlation coefficients among patients' profile entropy values and the values of the characteristics of patient heterogeneity.

| Characteristic of patient heterogeneity | Pearson correlation |
|---|---|
| # of all providers | 0.68 |
| # of medical providers | 0.68 |
| # of pharmacy providers | 0.69 |
| Average # of all claim events | 0.66 |
| # of medical claim events | 0.67 |
| # of pharmacy claim events | 0.66 |
| # of unique codes (all) | 0.70 |
| # of unique diagnosis codes | 0.69 |
| # of unique procedural codes | 0.72 |
| # of unique medication codes | 0.68 |

Table E2. MAPE improvement and link between data heterogeneity and patient heterogeneity based on different types of patient categorization.

| | | Cost levels | | Entropy levels | | Need levels | | Entropy levels | |
|---|---|---|---|---|---|---|---|---|---|
| | **Granularity** | **Low cost** | **High cost** | **Low entropy** | **High entropy** | **Low need** | **High need** | **Low entropy** | **High entropy** |
| # of instances | - | 105,996 | 5,557 | 105,996 | 5,557 | 62,470 | 49,083 | 62,470 | 49,083 |
| Population (%) | - | 95 | 5 | 95 | 5 | 56 | 44 | 56 | 44 |
| Average # of all providers | Day | 3.96 | 5.02 | 3.84 | 6.22 | 2.56 | 5.89 | 3.31 | 4.92 |
| | Week | 4.29 | 5.86 | 4.24 | 6.74 | 2.66 | 6.60 | 3.77 | 5.19 |
| | Month | 5.11 | 6.54 | 4.95 | 7.96 | 2.90 | 8.13 | 4.71 | 5.85 |
| Average # of all claim events | Month | 22.63 | 32.23 | 21.02 | 51.12 | 12.40 | 36.14 | 18.66 | 28.19 |
| Average # of unique clinical codes | Month | 17.54 | 36.77 | 17.46 | 41.24 | 9.73 | 30.25 | 12.54 | 26.68 |
| Average entropy | Day | 0.29 | 0.62 | 0.28 | 0.71 | 0.21 | 0.41 | 0.28 | 0.36 |
| | Week | 0.32 | 0.67 | 0.31 | 0.78 | 0.23 | 0.48 | 0.29 | 0.41 |
| | Month | 0.41 | 0.84 | 0.4 | 0.91 | 0.28 | 0.64 | 0.34 | 0.57 |
| **MAPE improvement** | | **14.91** | **28.15** | **14.65** | **32.54** | **8.11** | **24.98** | **10.82** | **21.23** |

Table E3. Ablation study for identifying specific contributions of various components to MAPE prediction performance over different entropy buckets.

| Learning | EmdApp | Att | All | Heterogeneity percentile buckets | | | | |
|---|---|---|---|---|---|---|---|---|
| | | | | **0-20** | **20-40** | **40-60** | **60-80** | **80-100** |
| Single channel | Trainable | N | 62.9 | 39.1 | 44.9 | 58.6 | 76.8 | 94.2 |
| | Trainable | Y | 59.2 | 38.2 | 43.8 | 53.1 | 71.5 | 89.7 |
| | Doc2Vec | N | 57.9 | 36.9 | 41.8 | 51.7 | 70.7 | 88.5 |
| | Doc2Vec | Y | 56.4 | 35.7 | 41.2 | 50.3 | 68.4 | 86.2 |
| Channel-wise | Trainable | N | 51.1 | 33.3 | 38.4 | 46.2 | 61.5 | 76.3 |
| | Trainable | Y | 49.0 | 32.9 | 37.9 | 44.4 | 57.9 | 71.8 |
| | Doc2Vec | N | 48.1 | 32.7 | 37.2 | 43.9 | 56.8 | 69.8 |
| | **Doc2Vec** | **Y** | **46.8** | **32.3** | **35.6** | **43.8** | **53.7** | **66.6** |

**EmdApp**: Embedding approach. **Att**: Attention model.

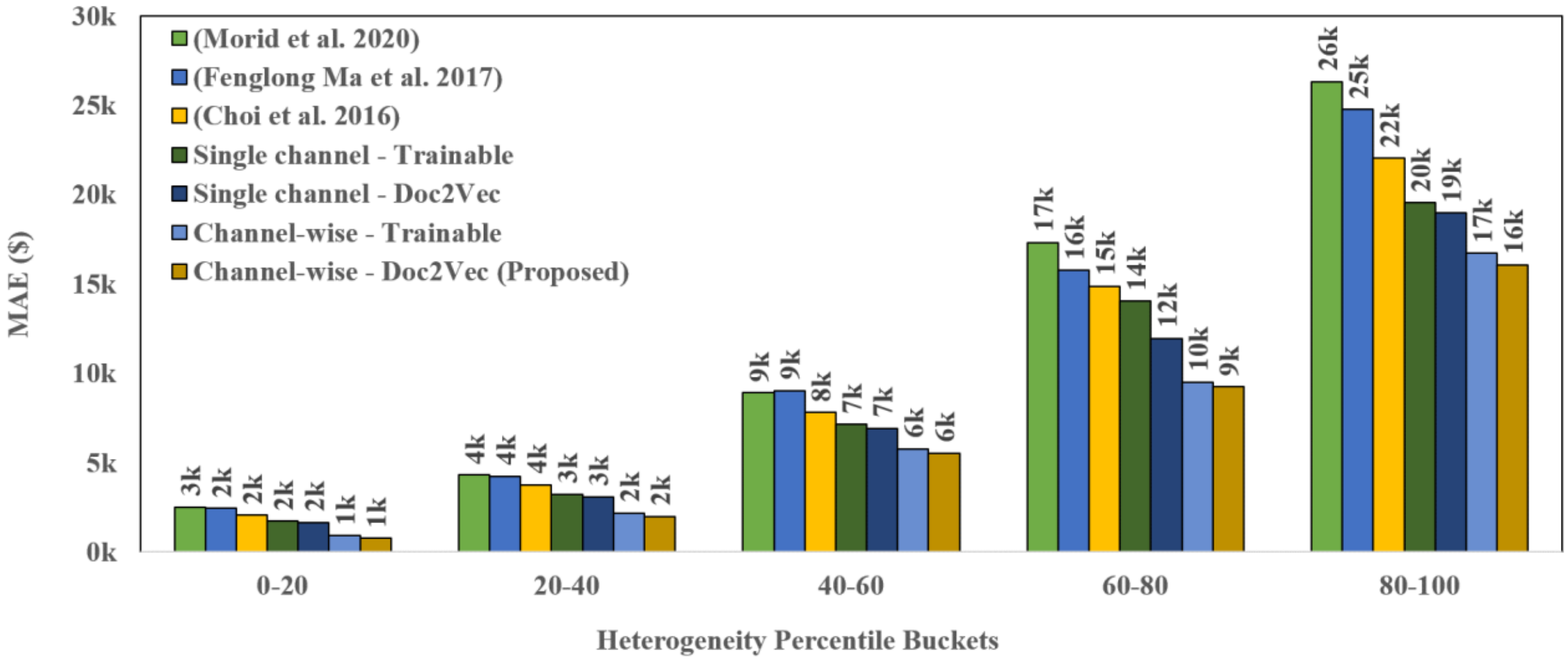

Figure E1. MAE for the dollar value of total claims cost prediction under different deep learning architectures over different entropy buckets.

# Appendix F: Benefits of the Proposed Architecture According to the Levels of Disparity and Heterogeneity Alignment

| *Data-individual Heterogeneity Alignment* | Low | High |
|---|---|---|
| **High** | **Q2: This study's benefits are higher than other cost-prediction models in Q2**<br>• The benefits of the study's architectural design could still be higher than other prediction models, justifying the adoption of the proposed channel-wise cost prediction design<br>• The uses and benefits of the proposed data heterogeneity measurement still apply | **Q1: This study's potential is the highest in Q1**<br>• Addressing data heterogeneity in representation learning reduces prediction errors and business outcome disparity<br>• A data heterogeneity measurement helps elucidate the alignment between patient and data heterogeneity<br>• This measurement facilitates flexible evaluations of the prediction models' ability to reduce business outcome disparity |
| **Low** | **Q4: The need to address both patient heterogeneity and data heterogeneity is less urgent in Q4**<br>• The potential benefits of the proposed architectural design and data heterogeneity measurement seem low | **Q3: Additional data preprocessing techniques could be beneficial in Q3**<br>• The benefit of reducing data heterogeneity during representation learning for predictions could be lower<br>• The value of using a data heterogeneity measurement to help evaluate the effect of prediction models on business outcome disparity for heterogeneous patients is limited<br>• Additional data collection, extraction, and augmentation designs could help reduce data heterogeneity during data pre-processing |

***Business Outcome Disparity*** (horizontal axis: **Low**, **High**)

Figure F1. The differential benefits of the proposed architecture and data heterogeneity metric

# Appendix G: Deep Learning Architectures for Patient Data

## EHR Deep Learning Architectures

Deep time series prediction architectures on EHR data can be categorized according to the nature of the input patient features (Morid et al. 2022). As shown in Figure G1.a, for numerical inputs such as lab tests and vital signs, the most common practice is to represent patients with a temporal matrix (Section 2.2) and then feed aggregated numerical input values for each time interval to a RNN model (Lipton et al. 2015, Purushotham et al. 2018, Sun et al. 2019, Zhang et al. 2017, 2019). These kinds of architectures must address challenges such as finding the appropriate windowing and imputing missing values (Lipton et al. 2016). Moreover, (Harutyunyan et al. 2019) proposed a *channel-wise* deep learning architecture on the MIMIC dataset, training a different RNN for each of the 17 numeric features, using a panoply of lab tests and recorded vital signs, including heart rate, body temperature, and blood pressure (see Figure G1.b). The underlying idea was that incorporating a separate RNN channel for each numeric feature aids in finding the missing patterns of individual features before combining them with the missing patterns of others.

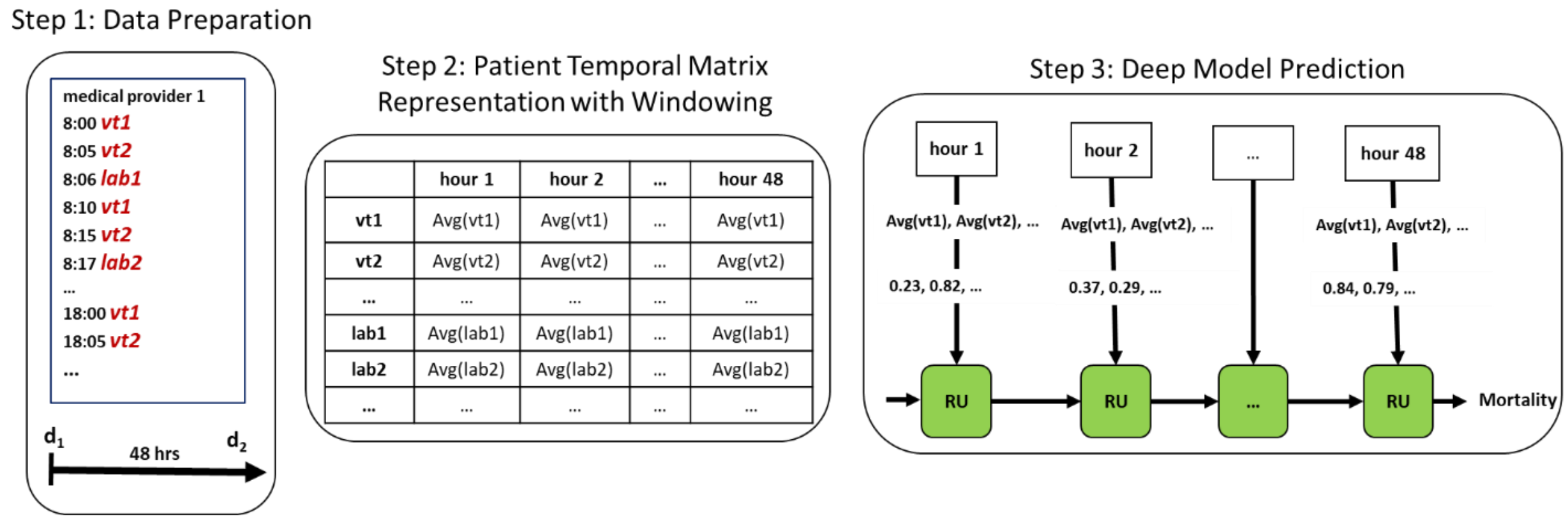

a.

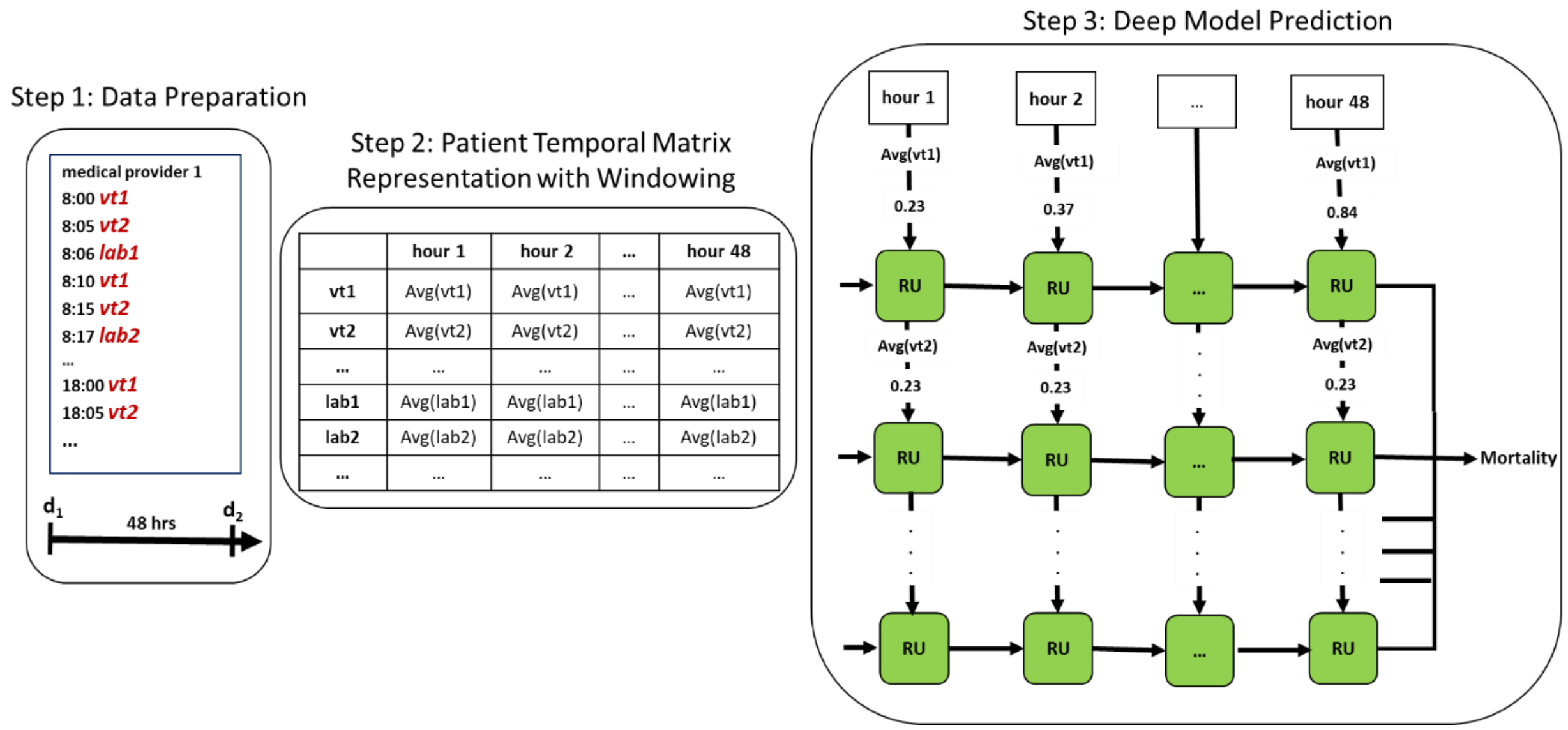

b.

Figure G1. EHR deep learning architectures with temporal-matrix representation for numerical data: a. Single channel and b. Multi-channel. Since mortality prediction using MIMIC is the most commonly used application for this type of architecture, the figure is constructed accordingly. Avg is a typical aggregation measure; vt1 and vt2 are examples of vital signs such as blood pressure, heart rate, or temperature; and lab1 and lab2 represent different lab tests such as sodium (Na), potassium (K), and white blood cell (WBC). All values are normalized as is common practice for RNNs. For the sake of simplicity, the recurrent memory units are generically shown as RU, without focusing on the specific RNN variant (LSTM, GRU, Bidirectional, etc.).

For categorical input features, the most common approach has an RNN model accept an input vector of embedded medical codes at each time step (Choi et al. 2017, 2018, Choi, Bahadori, Schuetz, et al. 2016, Gao et al. 2019). As shown in Figure G2.a., each code has its own timestamp and is embedded with pre-trained embedding using Skip-Gram (Section 2.2). This approach insists on each individual medical code having its own timestamp, since the order of the medical codes is important for both pre-trained embeddings and sequence representations. In contrast, the second approach has the RNN accept an input vector of embedded visits at each time step, where visits are embedded from unordered medical codes (Ayala Solares et al. 2020, Choi, Bahadori, Kulas, et al. 2016, Rebane et al. 2019). As shown in Figure G2.b., each visit has its own timestamp, and the embedding matrix is constructed during model training with an embedding layer (Section 2.2). This approach still requires the order of visits, available through EHR timestamps. Additionally, this method assumes that all medical codes occurring within a visit are related to each other, an acceptable assumption for EHR-visit data. A third alternative is to have an input vector of embedded visits at each time step, where each visit is the sum of the medical code embedding vectors within that visit (Choi, Schuetz, Stewart, et al. 2016). As shown in Figure G2.c., each code has its own timestamp and is embedded with pre-trained vectors using Skip-Gram (Section 2.2). This approach requires the timestamps of all medical codes and visits. Similar to the second input approach, summing up the medical codes inside each visit implicitly assumes that all medical codes therein are related to each other.

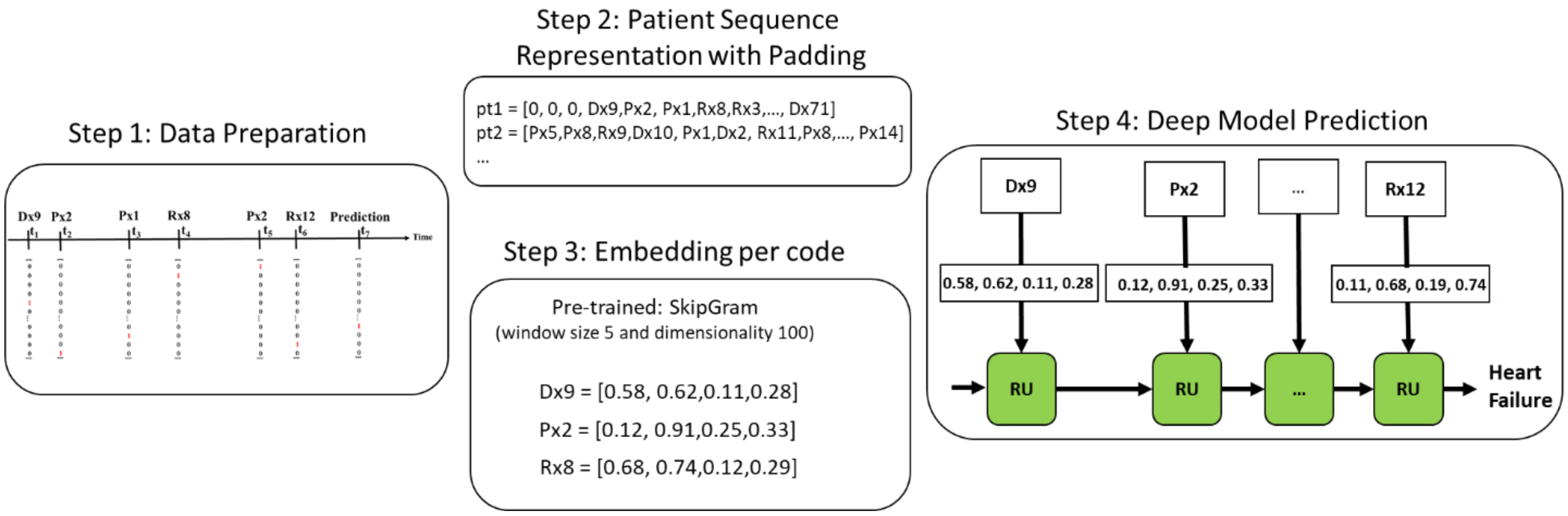

a.

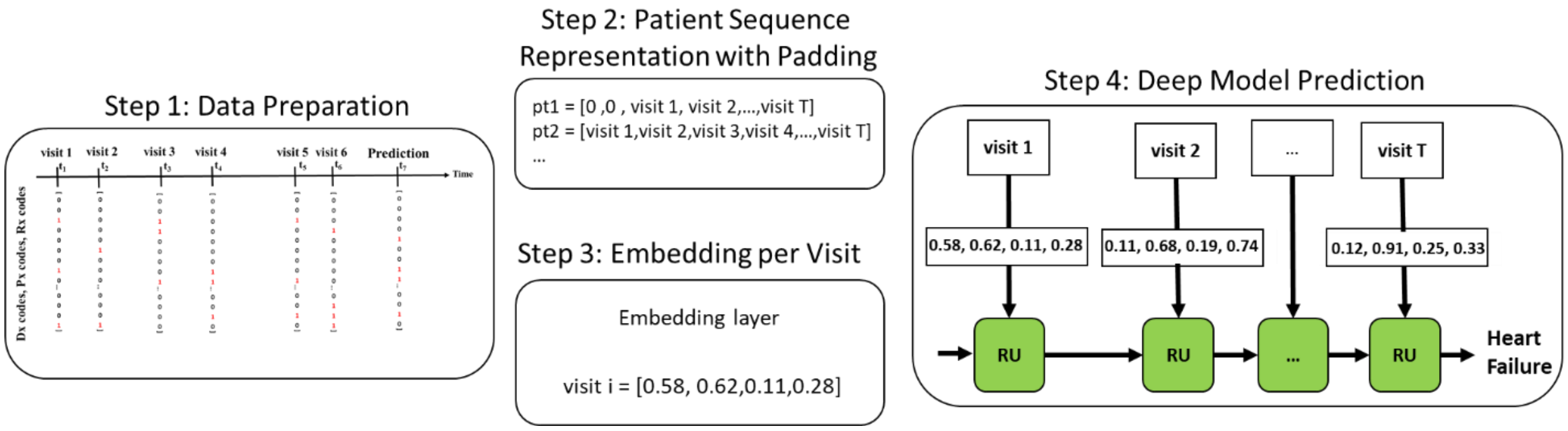

b.

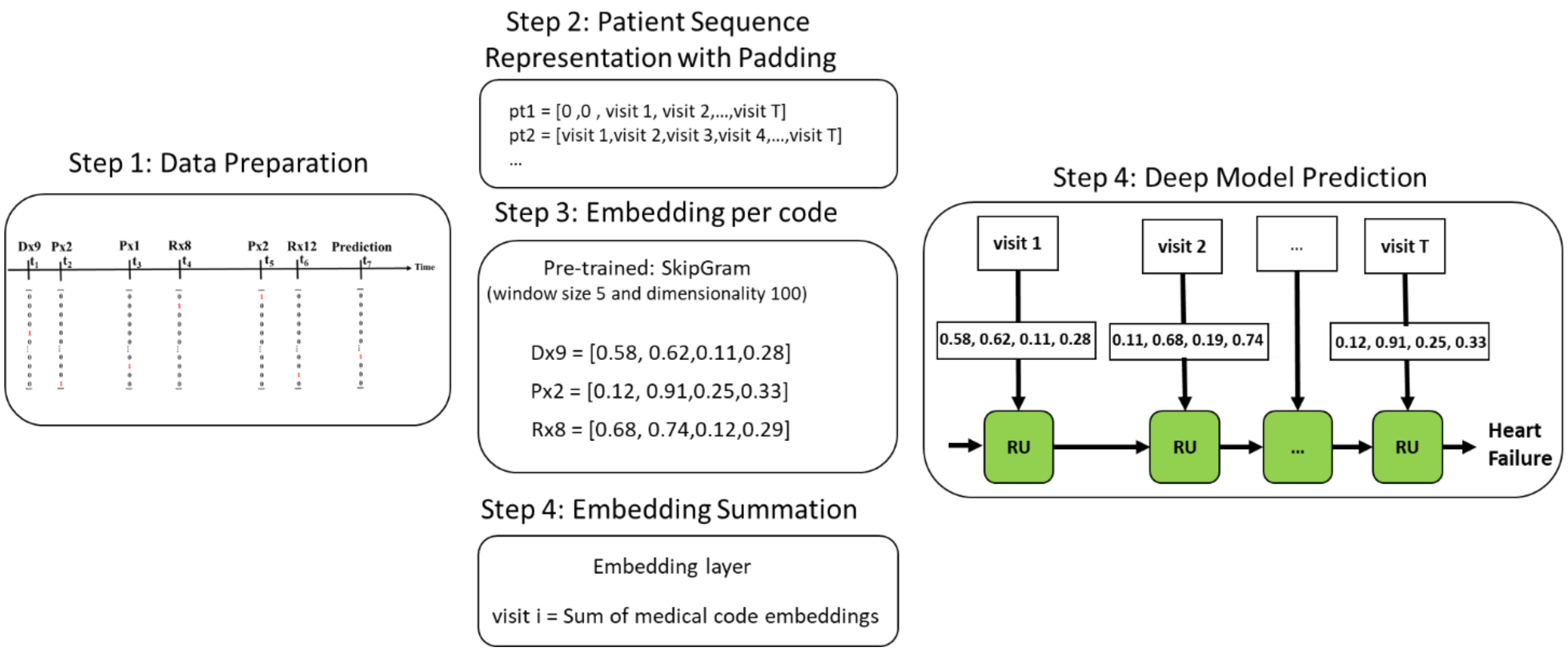

c.

Figure G2. EHR deep learning architectures with sequence representations for categorical data: a. Code embedding and code sequence representation, b. Visit embedding and visit sequence representation, and c. Code embedding and visit sequence representation. For the sake of simplicity, embeddings have a cardinality of four in all examples. Padding is applied on visit or code sequences to make all input patient profiles the same length. Again, all values are normalized, as is the most common practice for RNNs. For subfigures a and c, all codes with identical time stamps are ordered randomly.

When combining numerical inputs with categorical features, the most common approach is to convert numerical values to pertinent categorical representations, and then have an RNN model accept an input vector of embedded visits for each time step (Esteban et al. 2016, Ge et al. 2019, Lee and Hauskrecht 2019, Tomašev et al. 2019, Yu et al. 2020). To achieve this, numerical values have been labeled as *low*, *moderate*, *high*, *normal*, or *abnormal* based on medical expert domain knowledge (Morid et al. 2022). As shown in Figure G3.a, this architectural design is most similar to Figure G2.b, except that it includes this one additional preprocessing step. Besides depending on the order of the visits and counting on the dependency of the medical codes occurring within a visit, this approach differs in that it relies on expert domain knowledge for converting numerical values into categorical codes for input. Alternatively, one can render a unique categorical code for every existing combination of laboratory test and associated value within patient records, and then embed all codes accordingly (Rajkomar et al. 2018). Within hourly time windows, embeddings of all code types are aggregated by weighted averaging and then concatenated together. As shown in Figure G3.b, this architecture is most similar to Figure G2.c, except that it concatenates the averaged embeddings. Besides depending on the timestamp of each individual medical code (including the converted categorical codes of numerical values), averaging the embeddings within an hour assumes a strong dependency of the medical codes occurring within that time interval, which is a reasonable assumption for that time scale.

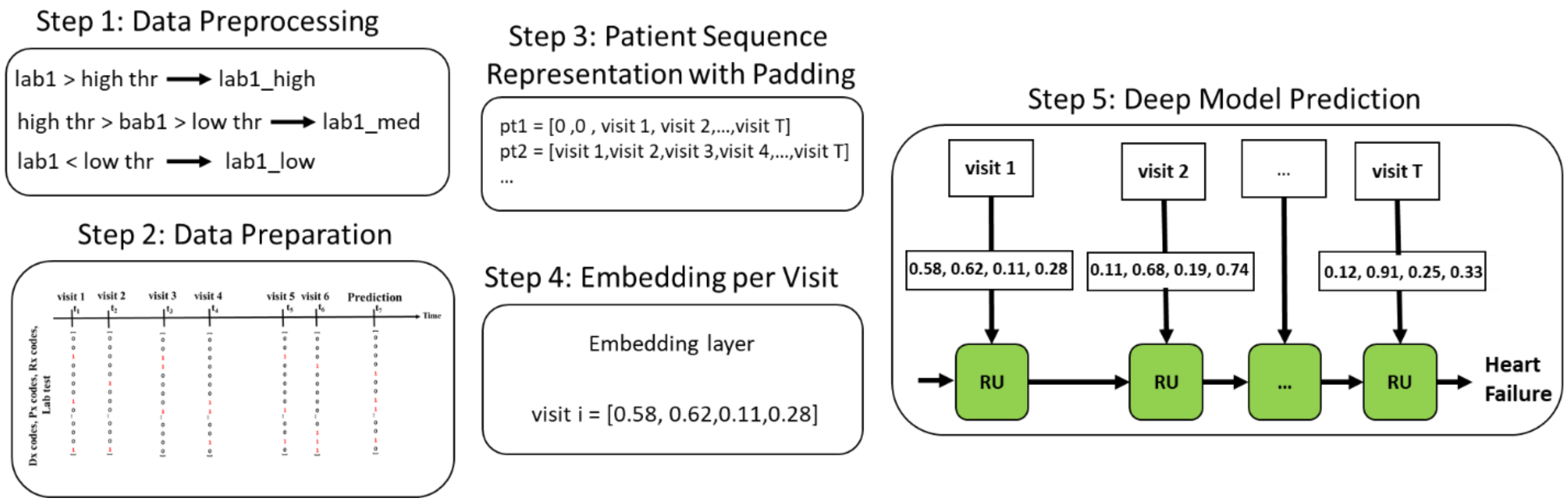

a.

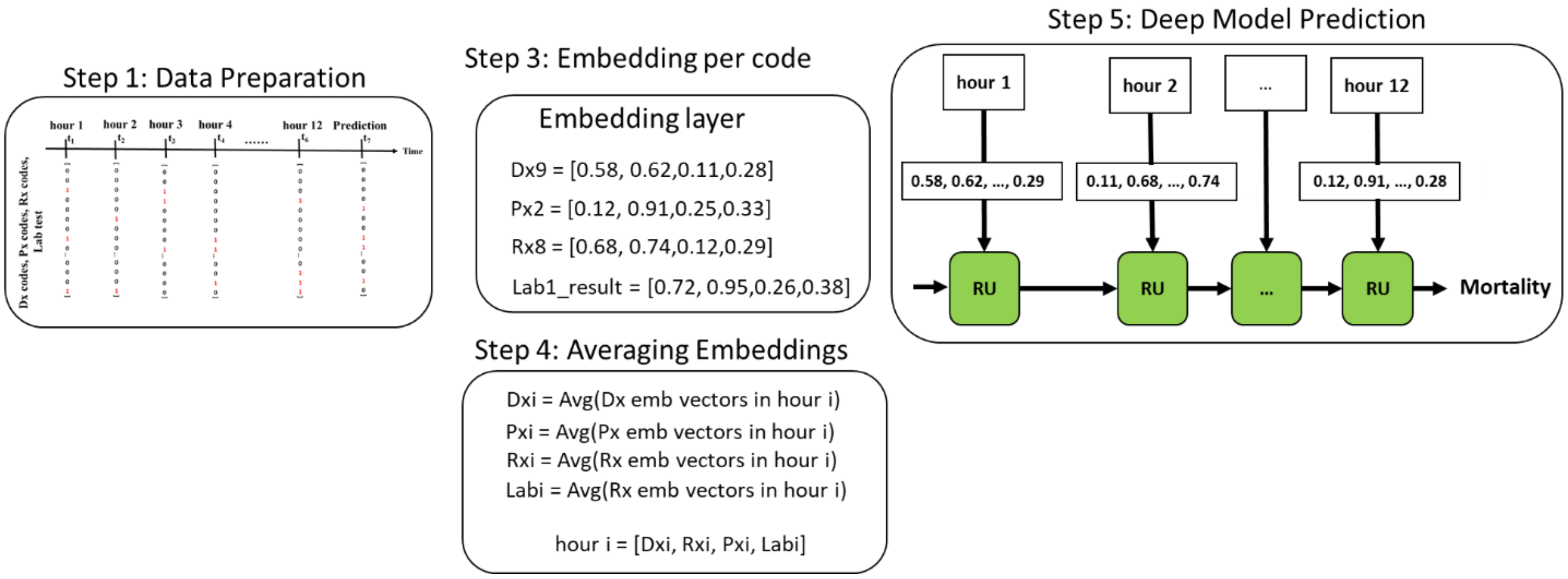

b.

Figure G3. EHR deep learning architectures with sequence representations for categorical and numerical data: a. Converting numerical values to categorical codes based on expert domain knowledge along with visit embeddings; b. Converting each individual numerical value to categorical codes, obtaining a weighted average of each code type, and concatenating the averaged embeddings per hour; and c. Code embedding and visit sequence representation. As in Figure G2, for the sake of simplicity, the embedding size is considered as four in all examples, padding is applied on visit sequences to make all patient profiles the same length, and all values are normalized.

While developing deep learning architectures for EHR has been extensively studied by deep learning researchers, the applicability of these architectures to AC data is not straightforward. Rather, it is important and necessary to ground the design-science study of cost prediction, specifically in regard to the way both the time series of medical codes and the charges of patient prior claims affect patients' future costs. Significantly, by utilizing the available timestamps of patient visits and medical codes, EHR deep learning studies are afforded meaningful relationships among codes associated with a single visit. However, AC data is aggregated daily, and neither the specific timestamps nor the guaranteed relationship is generally applicable. In fact, daily medical codes coming from multiple visits within a day, either dependent on each other or not, create a heterogeneous daily medical event that requires separate, deep learning architectures for discriminatory effect. Moreover, in the context of cost prediction, finding the optimal approach for combining categorical-code embedding vectors with numerical cost for AC architectures is not straightforward and requires in-depth experiments and analysis.

## AC Deep Learning Architectures

A recent systematic review on deep time series prediction in healthcare found that AC data has received significantly less attention than EHR data (Morid et al. 2022). Similar to EHR, deep time series prediction architectures on AC data can be categorized according to their patient representation approaches (Morid et al. 2022). As shown in Figure G4, the most common architecture uses temporal-matrix representations, usually with granularity at the monthly level and an aggregate of patient numerical or categorical features (Morid et al. 2020, Park et al. 2019). Here, medical codes are shown by their *count* per month (Park et al. 2019), and numerical features such as cost are shown by their *sum* over each month (Morid et al. 2020). While such architectures do not depend on the timestamp of medical codes and avoid the issues specified in Section 2.3.1, aggregating daily medical events in AC data up to a *month* naturally entails losing a significant amount of information that could be beneficial for model prediction.

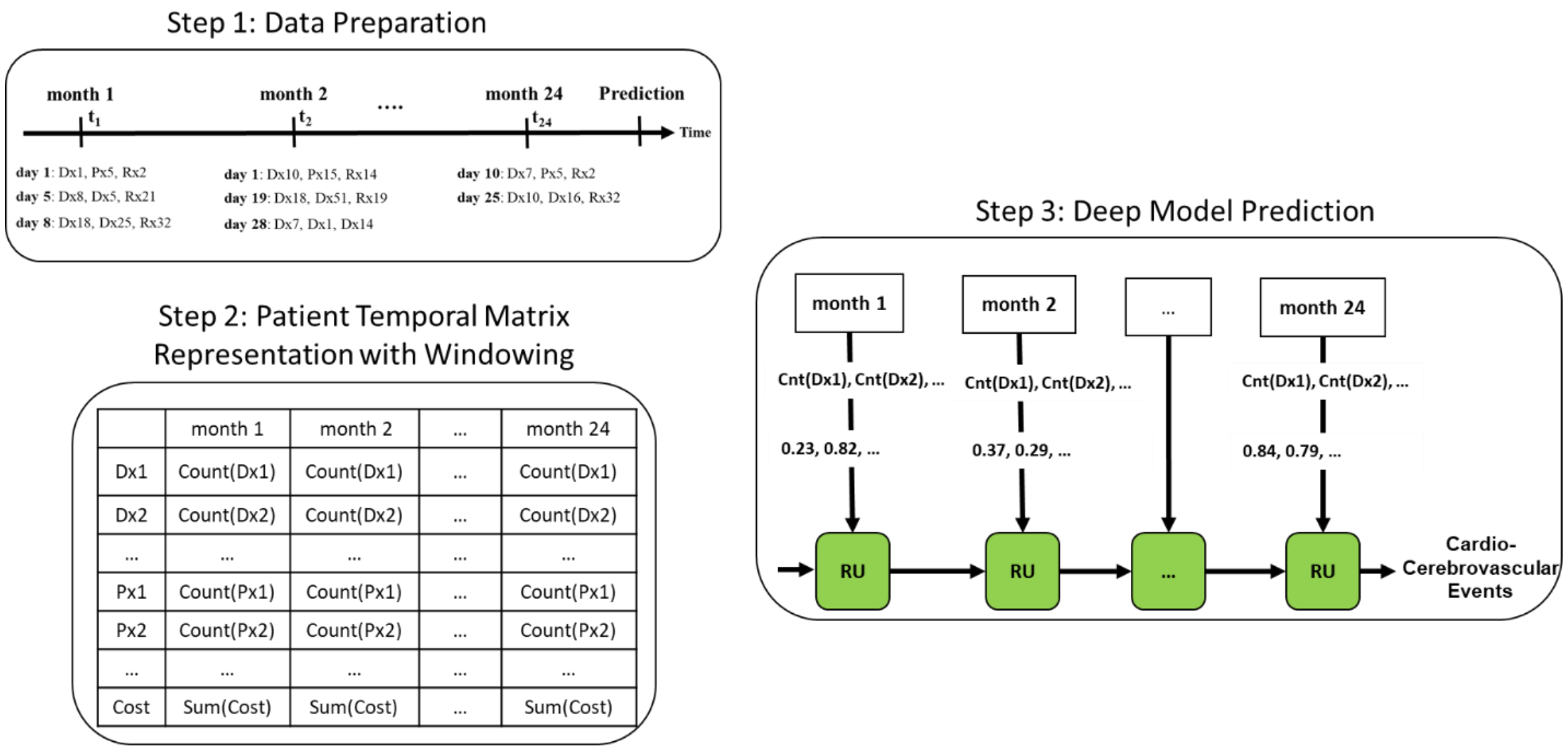

Figure G4. AC deep learning architecture with temporal-matrix representation for categorical and numerical data. For brevity, comments regarding normalization and memory cell units hold the same as in previous figures. Since the focus of this paper is on RNN, we have just shown RNN as the deep model architecture in the figure. Also, the input vector can include medical code groupings rather than individual codes, which is not shown in the figure to avoid unnecessary complexity.

When implementing sequence representation for AC data, one published approach embeds each medical code with a pre-trained embedding (Skip-Gram) and then represents each aggregated month of data with a weighted sum of all embeddings (Figure G5.a.) (Zeng et al. 2021). Here, same-day medical codes are randomly ordered for pre-trained embeddings. Each code has its own timestamp and is embedded with pre-trained Skip-Gram embedding (Section 2.2). Similar to the AC temporal-matrix architecture, with monthly data aggregation, this approach also neglects a significant amount of information that could be beneficial for the learning task. Moreover, summing up all embeddings into one vector may oversimplify the representation, leading to further information loss. A second approach for sequence representation with AC data is to have an RNN model that accepts an input vector of embedded weekly data at each time step, where the embeddings are aggregated from all unordered medical codes occurring in a given week (Fenglong Ma et al. 2017) (Figure G5.b.). While this second approach seems to be the most efficient method studied in the literature for AC data, it still suffers from the aforementioned information loss. Within a *week*, or even a *day*, a patient may incur several independent visits, and hence extracting the embeddings from multi-hot vectors of medical codes within that week may not be the most efficient approach for decreasing entropy. More importantly, the mixing of heterogeneous medical codes adds another layer of complexity to the representation learning step, which may eventually mislead the predictive model.

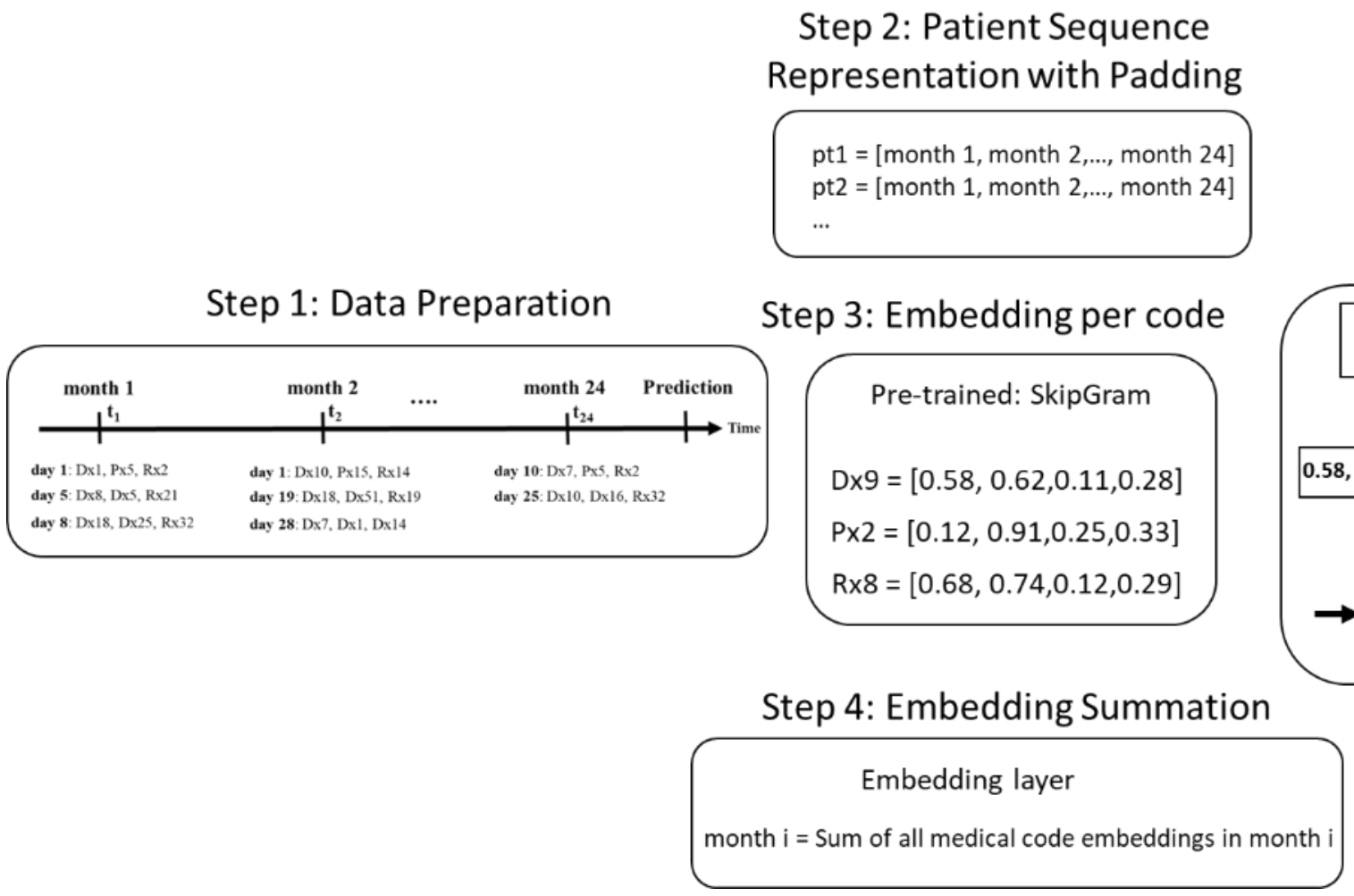

a.

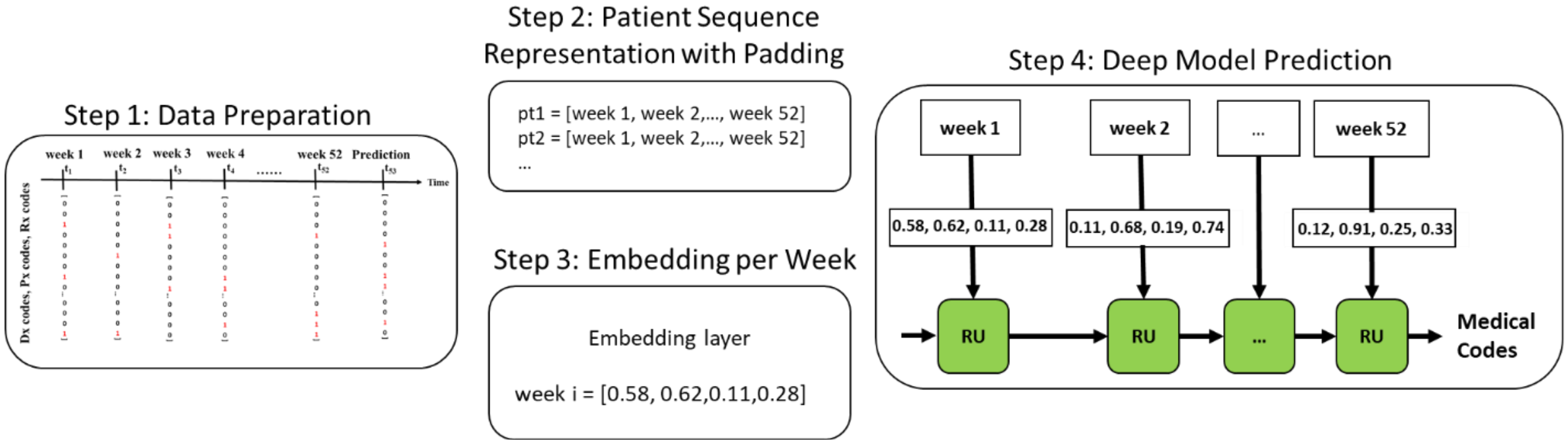

b.

Figure G5. AC deep learning architectures with sequence representations: a. Embedding medical codes with Skip-Gram and calculating the weighted sum of all codes within a month, and b. Embedding all weekly medical codes to represent each week by a single vector. For the sake of brevity, all qualifying statements regarding simplicity and context from Figure G4 still apply.

In this current research, we assert that, without any aggregation, the natural *daily* level granularity of AC data should be leveraged for accurate outcome prediction. This can be accomplished by a pre-trained embedding of daily medical events, motivated by EHR studies demonstrating that pre-trained embeddings of medical codes are more accurate than embeddings achieved with naïve embedding layers (Section 2.2). Second, heterogeneous medical codes occurring within a day should be separated by each type of medical code (diagnosis, procedure, or prescription) in order to reduce the complexity of temporal patterns for each type of code, and they should furthermore be extracted independently of each other.

# Appendix H: Multi-channel Entropy Examples

Table H1 provides examples of claim event entropies with various numbers of clinical codes. The multi-channel entropy index aims to capture the heterogeneity of patient profiles coming from the diversity of clinical codes and the frequency of claim events. The multi-channel part refers to multiple types of clinical codes, while the entropy part refers to the classic formula used to capture data heterogeneity (e.g., building decision trees).

Table H1. Examples of event entropies, where $E_{dx}$, $E_{px}$, and $E_{rx}$ are the total number of diagnosis codes, procedure codes and medication codes, respectively. Also, *Len (E)* is the total number of clinical codes, while *P(X)* and *Log(P(X))* respectively represent the probability and logarithmic probability value of each type of clinical code. In order to avoid *Log(0)*, all codes are increased by one: $P(E_{dx})= E_{dx} +1/Len(E)+3$, $P(E_{px})= E_{px} +1/Len(E)+3$, $P(E_{rx})= E_{rx} +1/Len(E)+3$.

| | Ex. 1 | Ex. 2 | Ex. 3 | Ex. 4 | Ex. 5 | Ex. 6 |
|---|---|---|---|---|---|---|
| **$E_{dx}$** | 12 | 10 | 4 | 6 | 4 | 2 |
| **$E_{px}$** | 0 | 1 | 4 | 0 | 1 | 2 |
| **$E_{rx}$** | 0 | 1 | 4 | 0 | 1 | 2 |
| **$P(E_{dx})$** | 0.867 | 0.733 | 0.333 | 0.778 | 0.556 | 0.333 |
| **$P(E_{px})$** | 0.067 | 0.133 | 0.333 | 0.111 | 0.222 | 0.333 |
| **$P(E_{rx})$** | 0.067 | 0.133 | 0.333 | 0.111 | 0.222 | 0.333 |
| **$log(P(E_{dx}))$** | -0.062 | -0.135 | -0.477 | -0.109 | -0.255 | -0.477 |
| **$log(P(E_{px}))$** | -1.176 | -0.875 | -0.477 | -0.954 | -0.653 | -0.477 |
| **$log(P(E_{rx}))$** | -1.176 | -0.875 | -0.477 | -0.954 | -0.653 | -0.477 |
| **$P(E_{dx}) * log(P(E_{dx}))$** | -0.054 | -0.099 | -0.159 | -0.085 | -0.142 | -0.159 |
| **$P(E_{px}) * log(P(E_{px}))$** | -0.078 | -0.117 | -0.159 | -0.106 | -0.145 | -0.159 |
| **$P(E_{rx}) * log(P(E_{rx}))$** | -0.078 | -0.117 | -0.159 | -0.106 | -0.145 | -0.159 |
| **$P(E_{dx}) * log(P(E_{dx})) + P(E_{px}) * log(P(E_{px})) + P(E_{rx}) * log(P(E_{rx}))$** | -0.211 | -0.332 | -0.477 | -0.297 | -0.432 | -0.477 |
| **Entropy** | -2.528 | -3.986 | -5.725 | -1.782 | -2.593 | -2.863 |

Ex.: Example